%% file: CellScientist_arXiv.tex
\documentclass{article}
\PassOptionsToPackage{table}{xcolor}
\PassOptionsToPackage{hyphens}{url}
\usepackage{cellscientist_arxiv,times}
\usepackage[utf8]{inputenc}
\usepackage[T1]{fontenc}
\usepackage{hyperref}
\hypersetup{hidelinks}
\usepackage{url}
\usepackage{xcolor}
\usepackage{graphicx}
\usepackage{caption}
\usepackage{nicefrac}
\usepackage{algorithm}
\usepackage{algorithmic}
\usepackage{newfloat}
\usepackage{listings}
\DeclareCaptionStyle{ruled}{labelfont=normalfont,labelsep=colon,strut=off}
\floatstyle{ruled}
\newfloat{listing}{tb}{lst}{}
\floatname{listing}{Listing}

\usepackage{booktabs}
\usepackage{subcaption}
\usepackage{cuted}
\usepackage{wrapfig}
\usepackage{placeins}
\usepackage{amsmath}
\usepackage{amssymb}
\usepackage{amsfonts}
\usepackage{mathtools}
\usepackage{amsthm}
\usepackage{bm}
\usepackage{microtype}
\usepackage{multirow}
\usepackage{pifont}
\usepackage{tabularx}
\usepackage{adjustbox}
\usepackage{tikz}
\usetikzlibrary{arrows.meta,positioning,shapes.geometric,fit,calc}
\usepackage{tcolorbox}
\tcbuselibrary{skins,breakable}

\definecolor{bg_code}{RGB}{242,244,247}
\definecolor{bg_review}{RGB}{255,248,240}
\definecolor{bg_metric}{RGB}{240,255,244}
\definecolor{bg_report}{RGB}{248,248,255}
\definecolor{key_color}{RGB}{0,51,102}
\definecolor{var_color}{RGB}{180,0,0}
\definecolor{statistical}{RGB}{255,240,240}
\definecolor{deepmodels}{RGB}{240,255,240}
\definecolor{generative}{RGB}{240,245,255}
\definecolor{agentic}{RGB}{250,240,255}
\definecolor{ours}{RGB}{255,255,220}
\definecolor{csaccent}{RGB}{239,244,251}

\lstdefinelanguage{promptyaml}{
  keywords={system,user},
  keywordstyle=\color{key_color}\bfseries,
  identifierstyle=\color{black},
  sensitive=false,
  comment=[l]{\#},
  commentstyle=\color{gray}\itshape,
  stringstyle=\color{black},
  morestring=[b]',
  morestring=[b]",
  moredelim=[s][\color{var_color}]{\$\{}{\}}
}

\providecommand{\cellres}[2]{}
\renewcommand{\cellres}[2]{\ensuremath{#1_{\scriptscriptstyle \pm #2}}}
\providecommand{\cellbest}[2]{}
\renewcommand{\cellbest}[2]{\ensuremath{\mathbf{#1}_{\scriptscriptstyle \pm #2}}}
\providecommand{\effectci}[3]{}
\renewcommand{\effectci}[3]{\ensuremath{#1\,\text{\tiny [#2,#3]}}}

\theoremstyle{plain}

\theoremstyle{definition}

\theoremstyle{remark}

\title{CellScientist: From Execution Feedback to Auditable Model-Revision Trajectories for Cellular Perturbation Prediction}
\author{%
  Mengran Li\textsuperscript{1,2,*}\quad
  Bo Li\textsuperscript{3,*}\quad
  Jiaying Wang\textsuperscript{2}\quad
  Wenbin Xing\textsuperscript{2}\quad
  Chengyang Zhang\textsuperscript{4}\\[2pt]
  Jinlin Wu\textsuperscript{1}\quad
  Zhen Lei\textsuperscript{1}\quad
  Jiebo Luo\textsuperscript{1}\quad
  Stan Z. Li\textsuperscript{5}\quad
  Zelin Zang\textsuperscript{1,\ensuremath{\dagger}}%
}
\affiliations{%
  \textsuperscript{1}Centre for Artificial Intelligence and Robotics,\\
  Hong Kong Institute of Science \& Innovation, Chinese Academy of Sciences\\
  \textsuperscript{2}Sun Yat-sen University\quad
  \textsuperscript{3}University of Macau\\
  \textsuperscript{4}Sichuan University\quad
  \textsuperscript{5}Westlake University%
}
\authornotes{\textsuperscript{*}Equal contribution.\qquad
\textsuperscript{\ensuremath{\dagger}}Corresponding author: Zelin Zang.}
\hypersetup{
  pdftitle={CellScientist: From Execution Feedback to Auditable Model-Revision Trajectories for Cellular Perturbation Prediction},
  pdfauthor={Mengran Li; Bo Li; Jiaying Wang; Wenbin Xing; Chengyang Zhang; Jinlin Wu; Zhen Lei; Jiebo Luo; Stan Z. Li; Zelin Zang},
  pdfsubject={Preprint},
  pdfkeywords={cellular perturbation prediction, model revision, large language models, reproducibility}
}
\begin{document}

\maketitle

\begin{abstract}
Cellular perturbation-response modeling requires coordinated choices of representations, fusion mechanisms, objectives, and training procedures. Large language models (LLMs) can propose executable candidates, but unconstrained revision can produce invalid implementations, change task semantics, or discard useful components. We present \textbf{CellScientist}, a protocol-constrained workflow that converts execution and validation feedback into auditable model-revision trajectories. It records design states and outcomes, routes discrepancies to specific components, and applies local revisions under a fixed task contract. A matched-budget study fixes the candidate language, predictor, fitting, and evaluator: structured revision finds better held-out predictors at small budgets under two LLM backbones. Operational audits link history to fewer repeated proposals, contract checks to contained violations, and discrepancy routing to targeted repairs. Refits of two frozen designs on an independently acquired cohort evaluate external predictive utility. Open-workflow trajectories retain improvements, regressions, and failures, while transcriptomic and single-cell searches extend application to additional response spaces. CellScientist produces both a selected predictor and an inspectable record of its development.
Project page: \url{https://limengran98.github.io/CellScientist/}.
\end{abstract}

\section{Introduction}

Cellular perturbation-response assays measure how chemical or genetic interventions reshape cellular states. Complementary readouts include Cell Painting morphology, transcriptomic profiles, and single-cell measurements~\citep{stirling2021cellprofiler,seal2025cell,subramanian2017high}. Predicting these responses can extend observed effects to unmeasured perturbations and contexts, supporting compound characterization, mechanism-of-action analysis, and virtual-cell modeling.

Generalization remains difficult because measured responses combine perturbation-induced biology with assay- and context-specific variation. Morphological profiles depend on plate and processing factors~\citep{arevalo2024evaluating}, while transcriptomic responses vary across batches and cellular contexts~\citep{zhou2025quantifying,zhang2024recovery}. Effective predictors therefore require coordinated choices of perturbation and context representations, dose conditioning, feature interactions, objectives, and training procedures. Revising one component can change the behavior of others, making feedback attribution central to model development.

These dependencies create a feedback-attribution problem. A poor validation outcome does not identify whether the next change should target conditioning, representation, fusion, or the objective. Execution failures raise a different question: whether a candidate can be evaluated under the intended protocol. A useful search controller must distinguish these signals before converting them into a concrete revision.

AutoML systems search predefined model and hyperparameter spaces~\citep{wang2021flaml,gijsbers2024amlb}, whereas LLM-based modeling agents can generate and revise executable programs~\citep{jiang2025aide,lu2024ai,tang2025cellforge}. This flexibility also requires control over how execution and validation outcomes guide subsequent changes. Revisions must preserve task semantics, target components implicated by the feedback, and retain the evidence behind each decision. We therefore ask how execution feedback can guide localized model-design revisions while preserving a fixed task contract.

\begin{figure}[!ht]
    \centering
    \includegraphics[width=0.99\textwidth]{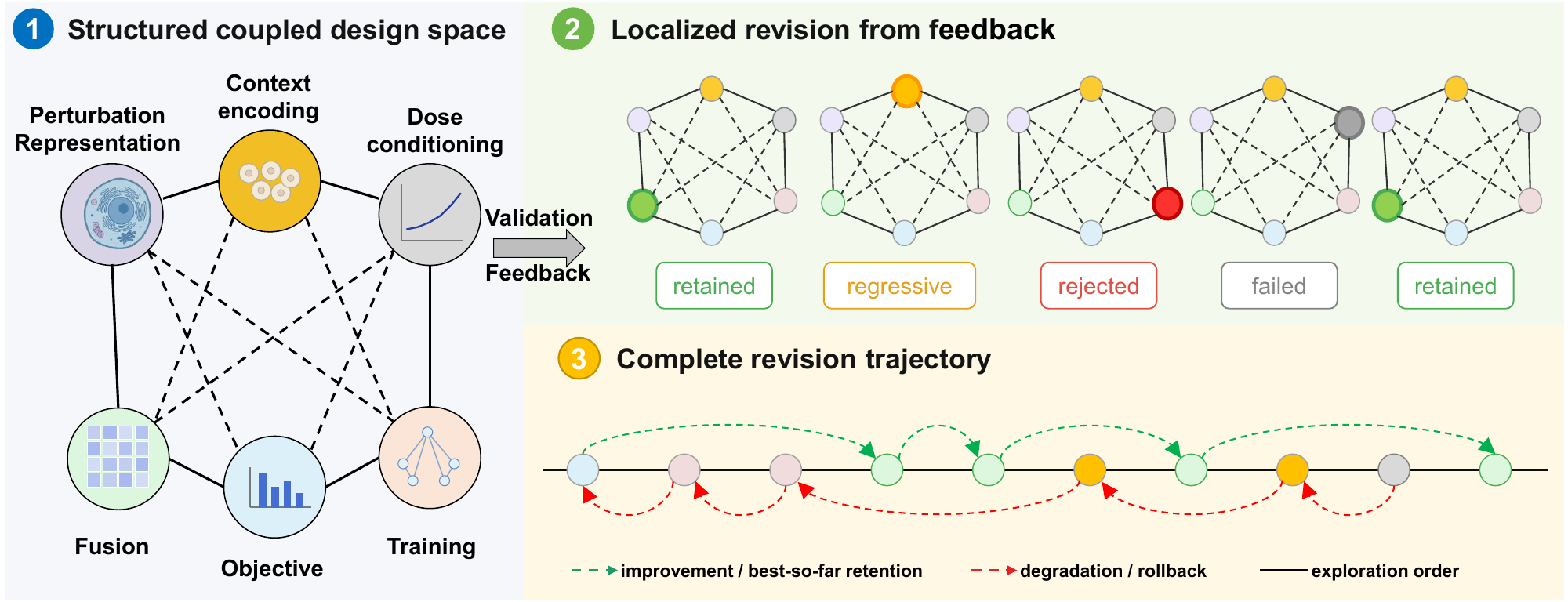}
    \caption{\textbf{Conceptual overview of CellScientist.} Revision operates over coupled choices of perturbation representation, context encoding, dose conditioning, fusion, objectives, and training strategy. Recorded feedback guides local updates, while the trajectory preserves improvements, regressions, rollbacks, rejected candidates, execution failures, and validation-retained states.}
    \label{fig:concept}
\end{figure}

We propose \textbf{CellScientist}, a framework for structured and auditable model revision in cellular perturbation-response prediction. It represents the coupled design space, routes feedback to implicated components, and records successive revisions (Figure~\ref{fig:concept}). Each state links a design, implementation, and outcome under a fixed task contract that preserves data, interface, and evaluation semantics.

\begin{figure}[!ht]
    \centering
    \includegraphics[width=0.95\textwidth]{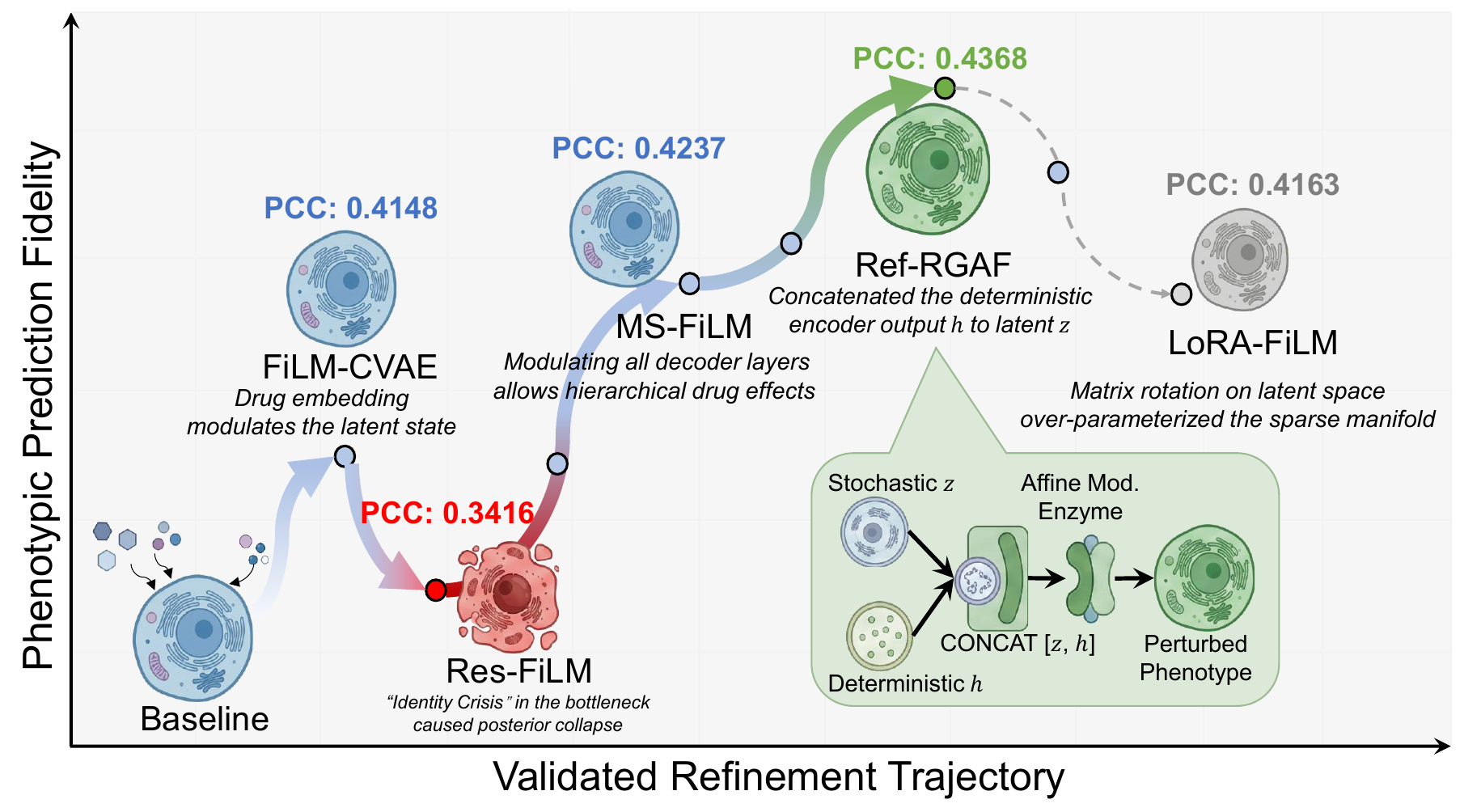}
    \caption{\textbf{Feedback-guided model-revision trajectory on BBBC047.} Recorded validation feedback guides successive revisions from modulation-based candidates to Ref-RGAF, which adds a deterministic control-reference encoding and attains the highest validation PCC in this trajectory. The annotations summarize working diagnostics recorded during the search. The retained record includes intermediate regressions and the later, lower-performing LoRA-FiLM candidate.}
    \label{fig:bbbc047_trajectory}
\end{figure}

Figure~\ref{fig:bbbc047_trajectory} illustrates this process on BBBC047. Modulation-based candidates show both improvements and regressions, prompting a working diagnosis that the supplied control-reference information is underused. A localized revision introduces a deterministic reference encoding alongside the stochastic latent state, yielding the validation-selected Ref-RGAF candidate. The recorded later regression makes clear why the selected model differs from the last attempted candidate. Appendix~\ref{app:case_b} gives the detailed trace.

CellScientist makes this transition explicit through three complementary functions. History-Aware Revision Tracking (HRT) maintains design states and evidence from previous attempts. Limited-Change Application (LCA) checks candidate eligibility and applies bounded repair. Performance-Discrepancy Refinement (PDR) links observed discrepancies to revision targets. Together, they connect the rationale for an edit to its implementation and subsequent validation, while preserving unsuccessful branches for inspection.

Our contributions are threefold. First, we formulate a \emph{model-design revision-routing problem} over coupled components under a fixed task contract. Second, CellScientist links local revisions to auditable design states, implementations, and outcomes. Third, matched-budget comparisons under two LLM backbones measure revision efficiency, operational audits examine the component roles, and external refits evaluate the predictive utility of frozen discovered designs.

\FloatBarrier

\section{Related Work}

\paragraph{Cellular perturbation-response modeling.}
Cell Painting models combine chemical structure, dose, and context to predict morphology~\citep{stirling2021cellprofiler,chandrasekaran2023jump,seal2025cell}. DeepCE, MultiDCP, and TranSiGen model expression~\citep{pham2021deepce,wu2022multidcp,tong2024transigen}; XPert uses biologically informed dual-branch modeling of cellular context and dose--time response~\citep{guo2026xpert}. GEARS predicts unseen genetic perturbations~\citep{roohani2024gears}, while MAP uses knowledge-aligned representations for unprofiled drugs~\citep{feng2026knowledge}. Systema separates perturbation-specific signals from systematic variation~\citep{vinas2026systema}. These approaches motivate representation, conditioning, and fusion as revision targets.

\paragraph{Representations and predictive evaluation.}
\citet{hasanaj2025multimodal} compare perturbation embeddings with a fixed downstream learner; MVCBench, CHMR, and MORSE examine representation quality and multimodal integration~\citep{li2026mvcbench,li2026chmr,zhang2026morse}. Tabular foundation models are competitive across cell-level, pseudobulk, and organismal perturbation tasks~\citep{palla2026tabular}. PerturBench highlights competitive simple models and perturbation-ranking metrics beyond reconstruction error~\citep{wu2025perturbench}. These findings motivate separating predictor quality from revision-policy efficiency: our controlled comparison fixes the candidate language, fitting procedure, and evaluator.

\paragraph{Automated model development.}
FLAML and AutoML benchmarks study predefined search spaces~\citep{wang2021flaml,gijsbers2024amlb}; AIDE and scientific agents extend proposals to executable programs~\citep{jiang2025aide,lu2024ai,turcan2025tusoai}. MLAgentBench and MLE-bench evaluate end-to-end experimentation~\citep{huang2024mlagentbench,chan2025mlebench}. MLE-STAR uses code-block ablations for targeted refinement~\citep{nam2025mlestar}. CellScientist evaluates diagnostic-ranked, history-filtered address routing under matched candidate budgets.

\paragraph{Biomedical agents and virtual cells.}
Virtual-cell LLM systems span direct predictors and tool-orchestrating agents~\citep{li2025llmsvirtualcell}. Biomedical agents address spatial analysis~\citep{wang2025spatialagent,yu2025gencellagent} and laboratory automation~\citep{qiu2025biomars,wang2025agentic}, with workflow evaluation studied by BioMLBench~\citep{miller2025biomlbench}. AssayBench evaluates gene-hit ranking from assay descriptions~\citep{debrouwer2026assaybench}. CellForge studies agent-designed perturbation models~\citep{tang2025cellforge}; VCWorld integrates biological knowledge~\citep{wei2026vcworld}; HarmonyCell combines hierarchical search, execution repair, and solution reuse~\citep{huang2026harmonycell}. CellScientist targets the revision decision within model development, linking feedback, component edits, and retained outcomes under protected task semantics.

\section{Methodology}
\label{sec:method}

\textbf{CellScientist} is a protocol-constrained controller for iterative perturbation-response model search. It links a model-design space $\mathcal{H}$ to an executable-candidate space $\mathcal{I}$ through the loop ``Design Specification $\rightarrow$ Executable Candidate $\rightarrow$ Validation Feedback $\rightarrow$ Local Revision,'' with $\mathbf{h}$ denoting a model-design state.

\begin{figure*}[ht]
    \centering
    \includegraphics[width=0.95\textwidth]{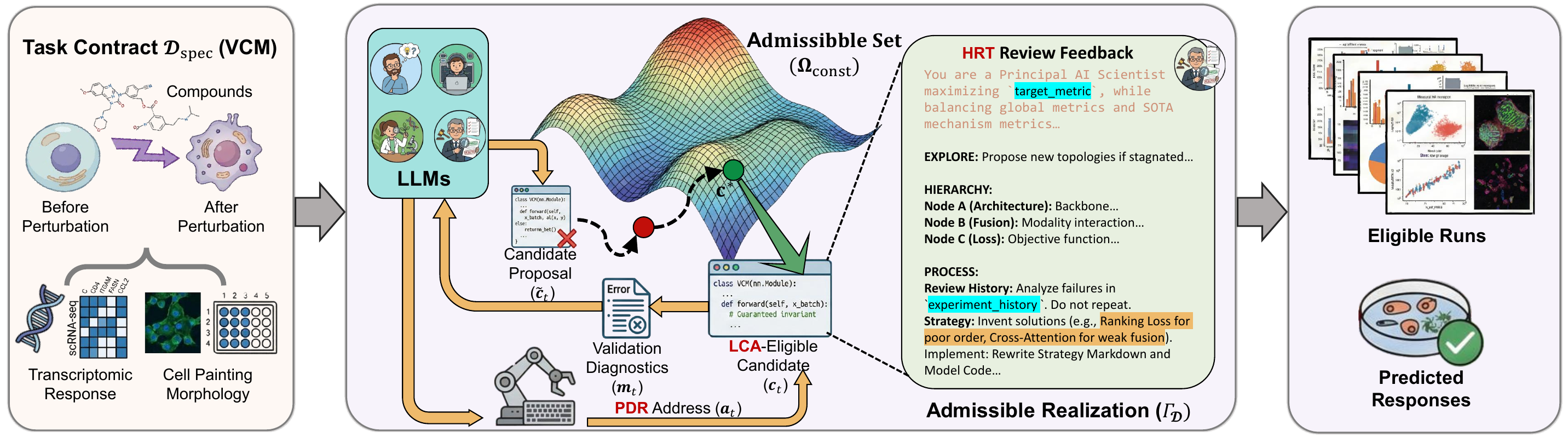}
\caption{
\textbf{Overview of the CellScientist framework.}
The task contract $\mathcal{D}_{\mathrm{spec}}$ fixes inputs, perturbation descriptors, response targets, data partitions, and evaluation semantics. HRT supplies the design state $\mathbf{h}_t$, task-state graph $\mathcal{G}_t$, and run history $\mathcal{B}_t$ for proposing $\tilde{\mathbf{c}}_t$. LCA applies configured checks and bounded repair to return an eligible candidate $\mathbf{c}_t$ or rejection $\bot$. Validation yields diagnostics $\mathbf{m}_t$, which PDR maps to a component address $a_t$ for localized revision. Validation selection retains the best eligible candidate; the complete trajectory is recorded.
}
    \label{fig:framework}
\end{figure*}

\subsection{Task Contract and Iterative Search}
\label{sec:dsbo}

Let $\mathbf{h}_t\in\mathcal{H}$ denote the model-design specification at iteration $t$ and let $\mathbf{c}_t\in\mathcal{I}$ denote an executable candidate. A task is specified by $\mathcal{D}_{\mathrm{spec}}=(\mathcal{X},\mathcal{P},\mathcal{Y},\mathcal{S},\mathcal{E})$, where $\mathcal{X}$ contains observed cellular context, $\mathcal{P}$ contains perturbation descriptors, $\mathcal{Y}$ contains response targets, $\mathcal{S}$ defines the data-partition protocol, and $\mathcal{E}$ defines the evaluation functional. The task contract remains fixed throughout a search trajectory.

The controller implements the following operational sequence:
\begin{equation}
\label{eq:dsbo}
\begin{aligned}
\tilde{\mathbf{c}}_t
&=\Phi(\mathbf{h}_t,\mathcal{G}_t,\mathcal{B}_t),\\
\mathbf{c}_t
&=\Gamma_{\mathcal{D}}(\tilde{\mathbf{c}}_t;\mathbf{h}_t,\mathcal{G}_t,\mathcal{B}_t,R)
\in\Omega_{\mathrm{const}}\cup\{\bot\},\\
\mathbf{m}_t
&=\mathcal{F}(\mathbf{c}_t;\mathcal{D}_{\mathrm{val}},\mathcal{E})\quad(\mathbf{c}_t\ne\bot),\\
\mathbf{h}_{t+1}
&=\mathbf{h}_t\oplus\mathcal{T}_{\mathrm{edit}}
\big(\mathcal{A}(\mathbf{m}_t,\mathcal{G}_t,\mathcal{B}_t)\mid\mathbf{m}_t\big).
\end{aligned}
\end{equation}
Here $\Phi$ requests candidate code or a registered candidate ID from the LLM, and $\Gamma_{\mathcal{D}}$ applies the configured checks and bounded repairs. For $\mathbf{c}_t=\bot$, $\mathbf{m}_t$ contains failure status and error text, with validation metrics marked unavailable. The state $\mathcal{Z}_t=(\mathbf{h}_t,\mathcal{G}_t,\mathcal{B}_t)$ contains the design, structured task record, and attempted implementations with outcomes. Section~\ref{sec:pdr} and Appendix~\ref{app:controlled_language} specify controlled routing and edit instantiation.

\subsection{HRT: History-Aware Revision Tracking}
\label{sec:hrt}

History-Aware Revision Tracking (HRT) represents the editable model-design state as a directed task-state graph $\mathcal{G}_t=(\mathcal{V}_t,\mathcal{E}_t)$. Nodes denote components such as perturbation representation, cellular-context encoding, fusion, architecture, objective design, and training strategy; edges record configured dependencies or ordering relations. The graph provides structured context for candidate generation and is updated with task status and evidence from prior attempts.

Given a diagnostic, the controller records intended revision targets and supplies their descriptions to proposal generation. Graph edges organize task context; the mutable-region registry and configured checks govern application. In the controlled study the graph records dependencies, while the typed grammar and history define legal one-field edits. HRT makes previous attempts available for filtering and retention.

\subsection{LCA: Limited-Change Application}
\label{sec:lca}

Limited-Change Application (LCA) governs candidate application and eligibility. Open-code proposals are applied to configured mutable notebook regions, followed by runtime checks and bounded repair. Controlled proposals undergo candidate-domain, allow-list, pairing, duplicate, prediction-shape, and finite-metric checks. Appendix~\ref{app:executable_checks} distinguishes these live checks from the separate registered contract-record audit. Failed attempts remain in the history and are excluded from selection.

Write $\chi_{\mathrm{I}}=1$ when the configured checks pass. For the applied candidate $\mathbf{c}^{(0)}$ and at most $R$ repair attempts, let $r^*$ be the first passing attempt. Then
\begin{equation}
\label{eq:lca_projection}
\Gamma_{\mathcal{D}}(\tilde{\mathbf{c}}_t;\mathbf{h}_t,\mathcal{G}_t,\mathcal{B}_t,R)
=\begin{cases}\mathbf{c}^{(r^*)},&\exists r\le R:\chi_{\mathrm{I}}(\mathbf{c}^{(r)})=1,\\
\bot,&\text{otherwise}.\end{cases}
\end{equation}
The passing set is $\Omega_{\mathrm{const}}=\{\mathbf{c}:\chi_{\mathrm{I}}(\mathbf{c})=1\}$. Passing candidates enter validation selection, and all outcomes are appended to $\mathcal{B}_{t+1}$.

\subsection{PDR: Performance-Discrepancy Refinement}
\label{sec:pdr}

Performance-Discrepancy Refinement (PDR) maps measured discrepancies to edit targets. Controlled feedback includes global PCC $r$, response PCC $r_\Delta$, top-50 response PCC $r_{\Delta,50}$, and error metrics. For feedback incumbent $(\bar{\mathbf h}_t,\bar{\mathbf m}_t)$, routing forms an ordered list $L_t$ and menu $Q_t$:
\begin{equation}
\label{eq:feedback_update}
\begin{aligned}
L_t&=\operatorname{FilterHistory}(\operatorname{Rank}(\bar{\mathbf m}_t,\bar{\mathbf h}_t,\mathcal{B}_t),\mathcal{B}_t),\\
Q_t&=\{(j,e_j(\bar{\mathbf h}_t,\mathcal{B}_t)):j\in\operatorname{FirstTwo}(L_t)\},\\
(a_t,\mathbf{h}_{t+1})&=\operatorname{LLMSelect}(Q_t\mid\bar{\mathbf m}_t,\bar{\mathbf h}_t).
\end{aligned}
\end{equation}
The five addresses are conditioning, target, input rank, output rank, and regularization. Fixed priorities first address missing conditioning, then direct targets or $r_\Delta-r>0.05$, then decoder discrepancy $r_{\Delta,50}<r_\Delta$, followed by capacity and fallback rules. History removes exhausted or unsuccessful addresses. Simultaneous diagnostics retain this order, with duplicates removed. Each $e_j$ selects one unseen one-field edit by the shared preference order; the LLM chooses an allowed pair, with the first pair as fallback. Appendix~\ref{app:controlled_language} specifies every rule and check.

\paragraph{Open-code revision.}
The open-code workflow uses a broader component vocabulary: the LLM proposes a focus and edits from diagnostics, graph context, and history; $\mathcal{T}_{\mathrm{edit}}$ records that revision request. An interface error can target an adapter, while suspected control-reference underuse can target an encoder or decoder. Appendix~\ref{app:protocol_controls} specifies this separate code-editing procedure.

In the BBBC047 example, the diagnosis becomes a decoder-input change: concatenate a deterministic reference encoding with the stochastic latent state. LCA checks the candidate, execution supplies its validation outcome, and HRT retains the edit and outcome, including the subsequent LoRA-FiLM regression. The record links a working diagnosis to an implementation change and the measured outcome used for selection.

\subsection{Trajectory Selection and Workflow}
\label{sec:workflow}

CellScientist stores each attempt in the run history. Let $\eta_t=\mathbb{I}\{\mathbf{c}_t\in\Omega_{\mathrm{const}}(\mathcal{D}_{\mathrm{spec}})\}$ indicate whether the proposed implementation passed the configured eligibility checks, and let $\mathcal{B}_{t+1}=\mathcal{B}_t\cup\{(\mathbf{h}_t,\tilde{\mathbf{c}}_t,\mathbf{c}_t,\mathbf{m}_t,a_t,\eta_t)\}$. The validation-selected candidate and its design state are given by
\begin{equation}
\label{eq:elite_selection}
(\mathbf{h}_{\mathrm{best}},\mathbf{c}_{\mathrm{best}})
=
\operatorname{Select}_{\mathrm{val}}
\left(
\{(\mathbf{h}_i,\mathbf{c}_i,\mathbf{m}_i):\eta_i=1\}_{i=0}^{T-1},
\mathcal{E}
\right),
\end{equation}
which returns the best eligible model-design state and executable candidate under the configured validation rule.

\section{Experiments}

The central empirical question is whether recorded feedback helps a controller spend a limited revision budget more effectively. We first compare controllers under a shared candidate language and held-out evaluator, then examine the operational roles of HRT, LCA, and PDR. Finally, we test two frozen model designs on an independently acquired cohort and connect model selection to the recorded open-workflow trajectories.

\textbf{RQ1: Revision efficiency.} Under a shared candidate language, initialization, fitting, and evaluation budget, does structured revision find better predictors earlier, and does the pattern recur with a second LLM backbone?
\textbf{RQ2: Operational mechanisms.} How do persistent history, contract checks, and discrepancy routing affect repeated proposals, invalid evaluations, and fault localization?
\textbf{RQ3: Predictive utility and revision records.} Do frozen discovered designs retain predictive utility after external refitting, and what do their full trajectories reveal about model selection?

\begin{table}[!t]
\centering
\caption{
\textbf{Five-seed matched-budget controller comparison on held-out global PCC.}
Upper: mean$\pm$sample SD over five paired controller/fitting seeds. Lower: CellScientist-minus-comparator effects with four-setting 95\% $t$ intervals, descriptive across the two cohorts and their grouping protocols. Policies share initialization, candidate language, fitting, and evaluation budget, including $h_0$. Blue identifies CellScientist; \textbf{bold} marks the best mean.
}
\label{tab:matched_pcc_effects}
\footnotesize
\renewcommand{\arraystretch}{1.25}
\setlength{\tabcolsep}{2.0pt}
\begin{tabular}{@{}llc|cccc>{\columncolor{csaccent}}c@{}}
\toprule
\textbf{Dataset} & \textbf{Split} & \textbf{Budget}
& \textbf{Fixed $h_0$}
& \textbf{Flat retry}
& \textbf{Random routing}
& \textbf{AIDE-style}
& \textbf{CellScientist} \\
\midrule
\multirow{3}{*}{BBBC036} & \multirow{3}{*}{Plate}
& 3  & \cellres{0.1664}{0.0005} & \cellres{0.1828}{0.0005} & \cellres{0.2250}{0.0493} & \cellres{0.2244}{0.0276} & \cellbest{0.2650}{0.0005} \\
& & 5  & \cellres{0.1664}{0.0005} & \cellres{0.2073}{0.0056} & \cellres{0.2277}{0.0502} & \cellres{0.2477}{0.0294} & \cellbest{0.2674}{0.0003} \\
& & 10 & \cellres{0.1664}{0.0005} & \cellres{0.2175}{0.0061} & \cellres{0.2626}{0.0077} & \cellres{0.2562}{0.0205} & \cellbest{0.2693}{0.0012} \\
\midrule
\multirow{3}{*}{BBBC036} & \multirow{3}{*}{SMILES}
& 3  & \cellres{0.1547}{0.0005} & \cellres{0.1583}{0.0046} & \cellres{0.2052}{0.0413} & \cellres{0.2058}{0.0230} & \cellbest{0.2336}{0.0016} \\
& & 5  & \cellres{0.1547}{0.0005} & \cellres{0.1882}{0.0129} & \cellres{0.2049}{0.0417} & \cellres{0.2231}{0.0169} & \cellbest{0.2346}{0.0009} \\
& & 10 & \cellres{0.1547}{0.0005} & \cellres{0.1948}{0.0005} & \cellbest{0.2347}{0.0016} & \cellres{0.2235}{0.0164} & \cellres{0.2346}{0.0009} \\
\midrule
\multirow{3}{*}{BBBC047} & \multirow{3}{*}{Plate}
& 3  & \cellres{0.2442}{0.0008} & \cellres{0.2500}{0.0055} & \cellres{0.3032}{0.0500} & \cellres{0.3013}{0.0381} & \cellbest{0.3396}{0.0004} \\
& & 5  & \cellres{0.2442}{0.0008} & \cellres{0.2630}{0.0130} & \cellres{0.3066}{0.0514} & \cellres{0.3153}{0.0399} & \cellbest{0.3457}{0.0004} \\
& & 10 & \cellres{0.2442}{0.0008} & \cellres{0.2777}{0.0005} & \cellres{0.3458}{0.0027} & \cellres{0.3191}{0.0372} & \cellbest{0.3481}{0.0009} \\
\midrule
\multirow{3}{*}{BBBC047} & \multirow{3}{*}{SMILES}
& 3  & \cellres{0.2504}{0.0004} & \cellres{0.2651}{0.0002} & \cellres{0.3131}{0.0536} & \cellres{0.3123}{0.0346} & \cellbest{0.3529}{0.0010} \\
& & 5  & \cellres{0.2504}{0.0004} & \cellres{0.2916}{0.0003} & \cellres{0.3207}{0.0582} & \cellres{0.3371}{0.0417} & \cellbest{0.3614}{0.0003} \\
& & 10 & \cellres{0.2504}{0.0004} & \cellres{0.2920}{0.0003} & \cellres{0.3668}{0.0073} & \cellres{0.3560}{0.0359} & \cellbest{0.3751}{0.0017} \\
\bottomrule
\end{tabular}
\par\smallskip
\renewcommand{\effectci}[3]{\shortstack{$#1$\\{\scriptsize $[#2,#3]$}}}
\begin{tabular}{@{}c>{\columncolor{csaccent}}c>{\columncolor{csaccent}}c>{\columncolor{csaccent}}c>{\columncolor{csaccent}}c@{}}
\toprule
\rowcolor{csaccent}
\multicolumn{5}{l}{\textit{CellScientist advantage: mean effect with paired-setting 95\% CI}} \\
\rowcolor{csaccent}
\multicolumn{1}{c}{\textbf{Budget}}
& \textbf{vs.\ Fixed $h_0$}
& \textbf{vs.\ Flat}
& \textbf{vs.\ Random}
& \textbf{vs.\ AIDE-style} \\
\rowcolor{csaccent}
\multicolumn{1}{c}{$B=3$}
& \effectci{+0.0939}{+0.0773}{+0.1104}
& \effectci{+0.0837}{+0.0734}{+0.0940}
& \effectci{+0.0361}{+0.0275}{+0.0448}
& \effectci{+0.0368}{+0.0270}{+0.0466} \\
\rowcolor{csaccent}
\multicolumn{1}{c}{$B=5$}
& \effectci{+0.0984}{+0.0774}{+0.1193}
& \effectci{+0.0648}{+0.0404}{+0.0892}
& \effectci{+0.0373}{+0.0292}{+0.0454}
& \effectci{+0.0215}{+0.0088}{+0.0342} \\
\rowcolor{csaccent}
\multicolumn{1}{c}{$B=10$}
& \effectci{+0.1029}{+0.0737}{+0.1320}
& \effectci{+0.0612}{+0.0306}{+0.0919}
& \effectci{+0.0043}{-0.0018}{+0.0104}
& \effectci{+0.0181}{+0.0052}{+0.0309} \\
\bottomrule
\end{tabular}
\end{table}

\subsection{Experimental Setup}
\label{sec:exp_setup}

\paragraph{Cellular response tasks.}
The main experiments use Cell Painting morphology profiles in CellProfiler feature space~\citep{stirling2021cellprofiler}. A predictor receives a vehicle-control reference, compound information, and dose, and predicts the treated morphology profile. BBBC036 and BBBC047~\citep{bray2017dataset} supply the controller comparison under plate-based and exact-SMILES-based grouping. LKCP Batch 2 supplies the external evaluation~\citep{natoli2021lincs,way2022morphology}. Global PCC is computed after flattening prediction and observation matrices; MSE and $R^2$ additionally measure prediction error and explained variation. Appendix~\ref{app:benchmarks} describes the cohorts and preprocessing.

\paragraph{Matched controller protocol.}
All policies share the initial state $h_0$, candidate language, low-rank predictor, fitting procedure, evaluator, paired seeds, and retention rule; revising policies also share the candidate budget. The language crosses conditioning (4 choices), response target (2), input rank (4), output rank (4), and regularization (6), yielding $4\times2\times4\times4\times6=768$ candidates. Their one-edit neighborhoods contain $3,1,3,3,5$ alternatives before history filtering. This universe is fixed for the reported comparison; per-step shortlists are policy-specific. CellScientist selects registered candidates through address routing. Appendix~\ref{app:controlled_language} specifies the domains, shared edit mapping, policy access, and fixed comparison protocol. Folds 1--3 fit candidates; disjoint fold-4 subsets provide feedback and selection scores; fold 5 reports held-out performance. Retention uses the mean feedback and selection PCC. Five paired controller/fitting seeds are evaluated at $B\in\{1,3,5,10\}$, including $h_0$. Each paired seed initializes the same randomized fit across policies. Within each LLM backbone, decoding settings are fixed.

\paragraph{External design protocol.}
The DD-MIA and Ref-RGAF architectures selected on BBBC036 and BBBC047 are frozen and fitted afresh to LKCP. Compound-disjoint partitions contain 33 training compounds, 15 validation compounds, and 16 test compounds. RealMLP~\citep{holzmuller2024better}, RF-SKL~\citep{breiman2001random,pedregosa2011scikit}, early-fusion MLP, and control-only MLP use the same target and split. Five neural refits assess training variation; RF-SKL uses one fixed-seed fit. Dataset construction, predictor settings, and training rules are given in Appendices~\ref{app:exp_morph}, \ref{app:morph_baselines}, and~\ref{app:implementation_config}.

\subsection{Matched-Budget Revision Efficiency (RQ1)}
\label{sec:rq1_budget_efficiency}

\paragraph{Matched revision policies.}
Fixed $h_0$ performs no revision. Flat retry selects among up to 12 unseen one-step neighbors using validation feedback but without a typed edit address. Random routing samples uniformly over all addresses with an unseen legal neighbor, followed by the same deterministic address-to-edit mapping used by CellScientist. The AIDE-style shared-language policy retains draft-then-greedy search~\citep{jiang2025aide}: after $h_0$, it allocates up to five evaluations to independent drafts and the remaining budget to improvement. Each revising policy evaluates exactly $B$ candidates. CellScientist ranks addresses using current diagnostics and HRT history, presents at most two eligible address--candidate pairs to Gemini 3 Pro, and applies LCA checks before evaluating the selected candidate. Capacity priorities use improvements over otherwise identical lower-rank candidates, following the rules in Appendix~\ref{app:controlled_language}. The comparison measures revision decisions within this shared executable language.

\paragraph{Useful candidates arrive earlier.}
Across the four task/split settings, CellScientist attains higher mean PCC than all comparison policies at $B=3$ and $B=5$ (Table~\ref{tab:matched_pcc_effects}). Against Random routing, the mean gains are $0.0361$ and $0.0373$. At $B=3$, CellScientist SDs range from $0.0004$ to $0.0016$, versus $0.0413$ to $0.0536$ for Random routing. These comparisons evaluate the combined diagnostic-priority, history-filtering, and LLM-selection policy within the fixed language.

At $B=10$, CellScientist exceeds Random routing in three settings and has a similar mean on BBBC036--SMILES ($0.2346$ versus $0.2347$). Its average advantage over Random routing narrows to $0.0043$, while the advantage over AIDE-style search remains $0.0181$ ($95\%$ CI $[0.0052,0.0309]$). Normalized frontier AUC summarizes predictive quality across the evaluation budget. CellScientist improves this measure over Fixed $h_0$, Flat retry, Random routing, and AIDE-style search by $0.0877$, $0.0608$, $0.0237$, and $0.0215$, with respective 95\% CIs $[0.0680,0.1074]$, $[0.0414,0.0802]$, $[0.0172,0.0303]$, and $[0.0117,0.0314]$. Together, the budget curve and endpoint results identify the clearest benefit as faster discovery of useful candidates.

\paragraph{The budget pattern recurs across backbones.}
Replacing Gemini 3 Pro with Qwen2.5-0.5B-Instruct preserves the early-budget pattern (Table~\ref{tab:backbone_frontier_summary}): gains are $0.0364$, $0.0341$, and $0.0012$ at $B=3,5,10$. Both backbones improve normalized frontier AUC. The four-setting intervals summarize configuration differences; Plate and SMILES reuse cohorts. Within-setting paired-seed intervals include zero at $B=3,5$ (Appendix~\ref{app:run_level_uncertainty}). Appendix~\ref{app:llm_backbones} records controller calls and fallback accounting.

\begin{table}[!ht]
\centering
\caption{\textbf{The early-budget advantage recurs with an open-weight controller.} CellScientist minus Random routing: mean effect and descriptive four-setting 95\% $t$ interval, with five paired runs per setting. AUC is normalized frontier AUC.}
\label{tab:backbone_frontier_summary}
\footnotesize
\renewcommand{\arraystretch}{1.25}
\setlength{\tabcolsep}{5pt}
\renewcommand{\effectci}[3]{\shortstack{$#1$\\{\scriptsize $[#2,#3]$}}}
\begin{tabular}{@{}lcccc@{}}
\toprule
\textbf{Controller backbone} & $\boldsymbol{B=3}$ & $\boldsymbol{B=5}$ & $\boldsymbol{B=10}$ & \textbf{AUC} \\
\midrule
\rowcolor{csaccent}
Gemini 3 Pro & \effectci{+0.0361}{+0.0275}{+0.0448} & \effectci{+0.0373}{+0.0292}{+0.0454} & \effectci{+0.0043}{-0.0018}{+0.0104} & \effectci{+0.0237}{+0.0172}{+0.0303} \\
\rowcolor{csaccent}
Qwen2.5-0.5B-Instruct & \effectci{+0.0364}{+0.0261}{+0.0467} & \effectci{+0.0341}{+0.0256}{+0.0426} & \effectci{+0.0012}{-0.0032}{+0.0056} & \effectci{+0.0217}{+0.0158}{+0.0275} \\
\bottomrule
\end{tabular}
\end{table}

\FloatBarrier
\subsection{Operational Roles of History, Checks, and Routing (RQ2)}
\label{sec:rq2_operational_roles}

The three components act at different points in a revision: HRT makes previous attempts available to proposal generation, LCA determines whether a candidate remains eligible, and PDR assigns a discrepancy to a component to edit. We examine each operation with its corresponding measured outcome (Table~\ref{tab:mechanism_audit_summary}).

\paragraph{History prevents repeated proposals.}
Twenty matched Gemini-controlled runs vary only persistent proposal history under the same trained-candidate budget. Without history, repeated IDs are recorded and replaced by legal unseen candidates before evaluation. HRT eliminates repeated proposals: all 170 proposals are unique, compared with 82 of 180 without history. This measures the use of prior attempts at proposal time while keeping the number of trained candidates fixed.

\paragraph{Contract checks protect evaluation.}
The LCA audit injects 40 registered contract-record violations alongside 20 legal controls. It repairs 15 records, blocks 15 at the pre-run gate, rejects 10 injected runtime-output records, and accepts all legal controls. Across these specified violation families, 30 requests are corrected or blocked before evaluation.

\paragraph{Discrepancy routing localizes repairs.}
The PDR audit uses five calibration cases and 15 held-out typed-state faults, with three variants per address and constructed metric records. Repair means restoring the registered configuration condition. Structured routing restores all 15 in one round, with no unrelated edits; Flat retry and Random routing restore 33.3\% and 60.0\%. Appendix~\ref{app:lca_pdr_audits} specifies the cases and repair criteria. Predictive revision efficiency is measured separately in RQ1.

\begin{table}[!ht]
\centering
\small
\captionsetup[subtable]{font=small,labelfont=bf,skip=3pt}
\setlength{\tabcolsep}{5pt}
\renewcommand{\arraystretch}{1.15}
\caption{
\textbf{Operational audits of HRT, LCA, and PDR.}
\textbf{(a)} HRT: 20 matched Gemini-controlled runs, 200 evaluations, and 170/180 proposal steps with/without history.
\textbf{(b)} LCA: 40 injected violations and 20 legal controls; 30 pre-dispatch interventions.
\textbf{(c)} PDR: 15 held-out typed-state faults; rounds are capped at the five-round budget. Best values are \textbf{bolded}.
}
\begin{subtable}[t]{0.48\textwidth}
\centering
\caption{HRT history.}
\label{tab:hrt_history_audit}
\begin{tabular}{@{}l>{\columncolor{csaccent}}cc@{}}
\toprule
\textbf{Measure} & \textbf{HRT} & \textbf{w/o Hist.} \\
\midrule
Unique proposals $\uparrow$ & \textbf{170} & 82 \\
Repeated proposals $\downarrow$ & \textbf{0} & 98 \\
Repeat rate $\downarrow$ & \textbf{0.0\%} & 54.4\% \\
Unique ratio $\uparrow$ & \textbf{100.0\%} & 45.6\% \\
\bottomrule
\end{tabular}
\end{subtable}\hfill
\begin{subtable}[t]{0.48\textwidth}
\centering
\caption{LCA outcomes.}
\label{tab:lca_audit_summary}
\begin{tabular}{@{}lc>{\columncolor{csaccent}}c@{}}
\toprule
\textbf{Outcome} & \textbf{Cases} & \textbf{Result} \\
\midrule
Repair & 15 & 15/15 repaired \\
Pre-run block & 15 & 15/15 blocked \\
Post-run block & 10 & 10/10 blocked \\
Legal controls & 20 & 20/20 accepted \\
\bottomrule
\end{tabular}
\end{subtable}\par\smallskip
\begin{subtable}[t]{0.94\textwidth}
\centering
\caption{PDR held-out results.}
\label{tab:pdr_audit_summary}
\begin{tabular}{@{}lcccc@{}}
\toprule
\textbf{Policy} & \textbf{Top-1} & \textbf{Repair} & \textbf{Edits} & \textbf{Rounds} \\
\midrule
Flat & 13.3\% & 33.3\% & 3.20 & 4.13 \\
Random & 6.7\% & 60.0\% & 3.27 & 3.87 \\
\rowcolor{csaccent}
\textbf{Structured} & \textbf{100\%} & \textbf{100\%} & \textbf{0.00} & \textbf{1.00} \\
\bottomrule
\end{tabular}
\end{subtable}
\label{tab:mechanism_audit_summary}
\end{table}

\subsection{Frozen Designs and Recorded Revision Trajectories (RQ3)}
\label{sec:rq3_predictive_utility}

\paragraph{External predictive utility.}
LKCP evaluates whether the frozen discovered architectures retain predictive utility after refitting on a separately acquired cohort with 1,671 morphology features. On its compound-disjoint test set, CS-model (BBBC047) attains the highest mean PCC among the evaluated predictors, $0.1766$ versus $0.1496$ for RealMLP. CS-model (BBBC036) attains the lowest mean MSE and highest mean $R^2$ (Table~\ref{tab:lkcp_external_transfer}). Both frozen architectures exceed the early-fusion and control-only MLPs in mean PCC. The designs trade correlation against absolute error; seed-matched differences from RealMLP have intervals spanning zero (Appendix~\ref{app:run_level_uncertainty}).

\begin{table}[!ht]
\centering
\caption{\textbf{External test of frozen CellScientist designs on LKCP Batch 2.}
Compound-disjoint test results on 16 compounds (432 conditions). Entries are
mean$\pm$sample SD over five refits; RF-SKL has one fixed-seed fit.
CS-model (BBBC036/047) denotes the frozen DD-MIA/Ref-RGAF design, respectively.
Best mean values among the evaluated methods are \textbf{bolded}.}
\label{tab:lkcp_external_transfer}
\small
\setlength{\tabcolsep}{7pt}
\renewcommand{\arraystretch}{1.08}
\begin{tabular}{@{}lccc@{}}
\toprule
\textbf{Method} & \textbf{PCC}$\uparrow$ & \textbf{MSE}$\downarrow$ & \textbf{$R^2$}$\uparrow$ \\
\midrule
\rowcolor{csaccent}
\textbf{CS-model (BBBC036)} & \cellres{0.1449}{0.0555} & \cellbest{4.4920}{0.0696} & \cellbest{-0.0080}{0.0082} \\
\rowcolor{csaccent}
\textbf{CS-model (BBBC047)} & \cellbest{0.1766}{0.0536} & \cellres{4.5615}{0.1420} & \cellres{-0.0561}{0.0424} \\
\midrule
RealMLP & \cellres{0.1496}{0.0582} & \cellres{4.5032}{0.1076} & \cellres{-0.0086}{0.0238} \\
RF-SKL & $0.0590$ & $4.8402$ & $-0.1034$ \\
Early-fusion MLP & \cellres{0.0860}{0.0020} & \cellres{4.5630}{0.0022} & \cellres{-0.0184}{0.0011} \\
Control-only MLP & \cellres{0.0852}{0.0025} & \cellres{4.5637}{0.0026} & \cellres{-0.0184}{0.0014} \\
\bottomrule
\end{tabular}
\end{table}

\begin{figure}[!ht]
\centering
\begin{minipage}[t]{0.98\textwidth}
    \centering
    \begin{minipage}[t]{0.485\linewidth}
        \centering
        \includegraphics[width=\linewidth]{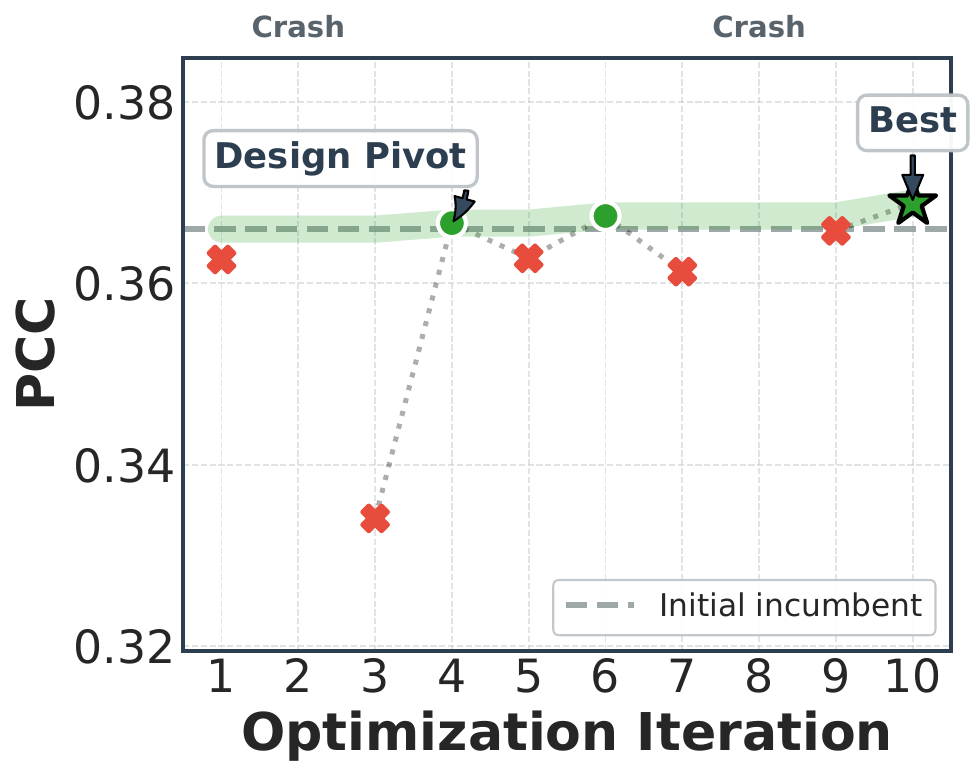}
        \par\footnotesize (a) BBBC036.
    \end{minipage}\hfill
    \begin{minipage}[t]{0.485\linewidth}
        \centering
        \includegraphics[width=\linewidth]{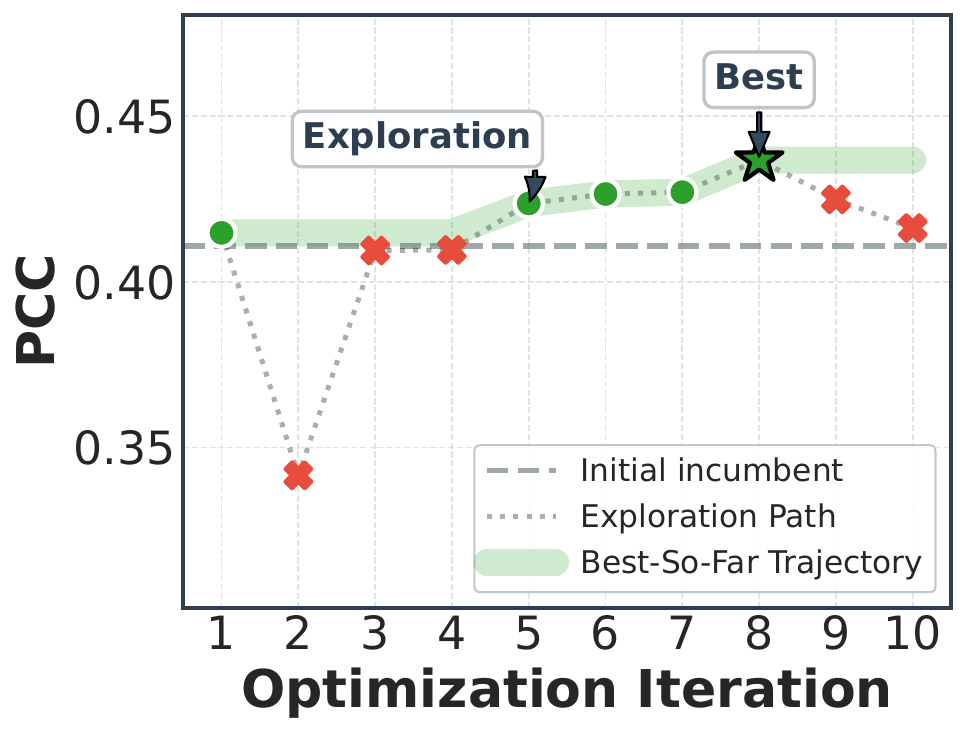}
        \par\footnotesize (b) BBBC047.
    \end{minipage}
    \captionof{figure}{\textbf{High-freedom workflow trajectories.} Valid candidate PCC and retained best-so-far over ten revisions. BBBC036 crashes at steps 2 and 8 yield no PCC; separate crash markers denote these failures, and retained best remains unchanged. Dashed references mark the incoming incumbent.}
    \label{fig:open_workflow_trajectory}
\end{minipage}
\end{figure}

\paragraph{From an attempted revision to a retained model.}
The open-workflow records show how these designs were reached (Figure~\ref{fig:open_workflow_trajectory}). On BBBC036, attempts 2 and 8 crash and produce no PCC; the retained incumbent remains available for subsequent revisions. Several executable candidates also regress before the trajectory reaches DD-MIA. On BBBC047, the revision to a deterministic control-reference encoding produces Ref-RGAF, whereas a later LoRA-FiLM attempt performs worse. Retaining the best eligible state therefore gives a different outcome from taking the last attempted model. Appendices~\ref{app:case_a} and~\ref{app:case_b} link these changes to the recorded diagnostics and implementations.

\FloatBarrier
The complete search outcomes are retained in Appendix~\ref{app:discovery_outcomes}: morphology validation results cover BBBC036, BBBC047, and CPG0016, and independent task-specific searches cover LINCS2020 expression prediction and the Norman and Papalexi single-cell readouts. Appendix~\ref{app:open_workflow_characterization} additionally records execution success, rule-compliant success, and runtime across cumulative workflow configurations. Together, these records document how the workflow proposes, evaluates, and retains models across cellular response tasks.

\section{Conclusion}

CellScientist converts execution and validation feedback into local model revisions under a fixed task contract. In the shared morphology candidate language, structured routing finds useful candidates earlier under two LLM backbones, highlighting the value of directing limited revision budgets toward components implicated by observed discrepancies. Operational audits further characterize the distinct roles of history, contract checks, and discrepancy routing by measuring repeated proposals, contract-record decisions, and registered fault repairs. External refits compare the predictive utility of two frozen designs on an independently acquired cohort, while complete revision trajectories retain failed, regressive, and validation-retained attempts behind model selection. Together, these results show how feedback-guided model development can produce not only an executable predictor, but also an inspectable record linking design decisions, implementation changes, and observed outcomes. Broader candidate languages, richer diagnostic signals, and cross-task reuse of revision knowledge are natural extensions of this formulation.

\clearpage
\subsection*{Reproducibility statement}
The operational algorithm and notation are provided in Appendix~\ref{app:method_algorithm_notation}. Dataset definitions, baseline protocols, evaluation metrics, implementation settings, and controller configurations are detailed in Appendices~\ref{app:benchmarks}--\ref{app:implementation_details}. The case studies retain failed and regressive candidates alongside validation-selected outcomes, while the registered LCA and held-out PDR audits are documented in Appendix~\ref{app:lca_pdr_audits}.

\bibliographystyle{iclr2027_conference}
\bibliography{references}

\clearpage
\appendix
\setcounter{secnumdepth}{3}
\raggedbottom
\makeatletter
\setlength{\@fptop}{0pt}
\setlength{\@fpsep}{14pt}
\makeatother

\section{Algorithm and Notation}
\label{app:method_algorithm_notation}

\textbf{Coupled model-design revision.}
Perturbation-response models combine interdependent choices in perturbation representation, cellular-context encoding, dose conditioning, feature interaction, objective design, and optimization. Changing one component can alter the behavior of others. Model revision therefore requires associating observed outcomes with candidate explanations and possible edit targets. An explicit record of component dependencies, accumulated evidence, and edit scope guides this process.

\textbf{Hypothesis--implementation alignment.}
Evaluating a model-design hypothesis requires an executable implementation that preserves the task's data, prediction target, interface, and evaluation semantics. Execution and validation outcomes then inform which design assumptions and components merit reconsideration. We formulate this as a \emph{model-design revision-routing problem}: mapping observed discrepancies to local design revisions while maintaining a fixed task contract.

\begin{algorithm*}[ht]
\caption{CellScientist Workflow}
\label{alg:cellscientist}
\begin{algorithmic}[1]
\REQUIRE Fixed task specification $\mathcal{D}_{\mathrm{spec}}$, initial model-design state $\mathbf{h}_0$, task-state graph $\mathcal{G}_0$, iteration budget $T$, repair budget $R$
\STATE Initialize run history $\mathcal{B}_0\leftarrow\emptyset$ and best eligible state $(\mathbf{h}_{\mathrm{best}},\mathbf{c}_{\mathrm{best}})\leftarrow(\mathbf{h}_0,\bot)$
\FOR{$t=0$ to $T-1$}
    \STATE Summarize the current model-design state, task-state graph, and run history
    \STATE Request a focus area and candidate edits $\tilde{\mathbf{c}}_t\leftarrow\Phi(\mathbf{h}_t,\mathcal{G}_t,\mathcal{B}_t)$
    \STATE Apply edits and configured checks with at most $R$ repair attempts: $\mathbf{c}_t\leftarrow\Gamma_{\mathcal{D}}(\tilde{\mathbf{c}}_t;\mathbf{h}_t,\mathcal{G}_t,\mathcal{B}_t,R)$
    \IF{$\mathbf{c}_t\neq\bot$}
        \STATE Execute $\mathbf{c}_t$ under the validation view of $\mathcal{D}_{\mathrm{spec}}$
        \STATE Record execution status, configured rule checks, and validation feedback $\mathbf{m}_t$
        \STATE Set $\eta_t\leftarrow 1$
    \ELSE
        \STATE Record execution or required-output failure in $\mathbf{m}_t$ and set $\eta_t\leftarrow 0$
    \ENDIF
    \STATE Route the diagnostic to an edit address $a_t\leftarrow\mathcal{A}(\mathbf{m}_t,\mathcal{G}_t,\mathcal{B}_t)$
    \STATE Form the next design record $\mathbf{h}_{t+1}\leftarrow\mathbf{h}_t\oplus\mathcal{T}_{\mathrm{edit}}(a_t\mid\mathbf{m}_t)$
    \STATE Append the addressed evidence and status to obtain $\mathcal{G}_{t+1}$
    \STATE Update run history $\mathcal{B}_{t+1}\leftarrow\mathcal{B}_t\cup\{(\mathbf{h}_t,\tilde{\mathbf{c}}_t,\mathbf{c}_t,\mathbf{m}_t,a_t,\eta_t)\}$
    \IF{$\eta_t=1$ and $\mathbf{c}_t$ improves the configured validation score}
        \STATE $(\mathbf{h}_{\mathrm{best}},\mathbf{c}_{\mathrm{best}})\leftarrow(\mathbf{h}_t,\mathbf{c}_t)$
    \ENDIF
\ENDFOR
\STATE \textbf{return} $(\mathbf{h}_{\mathrm{best}},\mathbf{c}_{\mathrm{best}})$ and the recorded trajectory $(\mathcal{G}_T,\mathcal{B}_T)$
\end{algorithmic}
\end{algorithm*}

\begin{table*}[t]
\centering
\caption{\textbf{Notation used in CellScientist.}}
\label{tab:notation}
\small
\renewcommand{\arraystretch}{1.02}
\setlength{\tabcolsep}{4pt}
\begin{tabularx}{\textwidth}{@{}>{\raggedright\arraybackslash}p{0.19\textwidth}X@{}}
\toprule
\textbf{Symbol} & \textbf{Description} \\
\midrule
$\mathcal{H}$ & Model-design space containing task-specific representation, fusion, objective, and training choices \\
$\mathcal{I}$ & Space of executable candidate programs \\
$\mathcal{D}_{\mathrm{spec}}$ & Fixed task specification $(\mathcal{X},\mathcal{P},\mathcal{Y},\mathcal{S},\mathcal{E})$ \\
$\mathcal{X},\mathcal{P},\mathcal{Y}$ & Cellular inputs, perturbation descriptors, and response targets \\
$\mathcal{S}$ & Data-partition protocol used by the configured task \\
$\mathcal{E}$ & Evaluation functional used for candidate scoring and selection \\
$\mathcal{D}_{\mathrm{train}},\mathcal{D}_{\mathrm{val}},\mathcal{D}_{\mathrm{test}}$ & Training, controller-feedback/selection, and reporting views induced by $\mathcal{S}$ \\
$\mathbf{h}_t$ & Model-design record at iteration $t$ \\
$\mathcal{G}_t=(\mathcal{V}_t,\mathcal{E}_t)$ & Structured task-state graph used for routing context and audit \\
$\mathcal{V}_t,\mathcal{E}_t$ & Task nodes and directed dependency edges \\
$\mathcal{B}_t$ & Run history storing attempted designs, implementations, and outcomes \\
$\mathcal{Z}_t=(\mathbf{h}_t,\mathcal{G}_t,\mathcal{B}_t)$ & Complete controller state at iteration $t$ \\
$T,R$ & Candidate-iteration budget and bounded LCA repair budget \\
$\Phi$ & LLM-conditioned candidate-proposal operator \\
$\Gamma_{\mathcal{D}}$ & Code-application, configured-check, bounded-repair, and rejection procedure \\
$\tilde{\mathbf{c}}_t$ & Proposed candidate edits before execution checks \\
$\mathbf{c}_t$ & Candidate produced after code application and checks \\
$\bot$ & Candidate that failed configured execution or required-output checks \\
$\eta_t$ & Indicator that $\mathbf{c}_t$ passed the configured eligibility checks \\
$\Omega_{\mathrm{const}}$ & Candidates that pass the implemented checks under $\mathcal{D}_{\mathrm{spec}}$ \\
$\chi_{\mathrm{I}},\mathcal{C}_{\mathrm{I}}$ & Eligibility indicator and its configured implementation-rule set \\
$\mathcal{F}$ & Feedback extraction operator \\
$\mathbf{m}_t$ & Structured feedback vector at iteration $t$ \\
$\bar{\mathbf h}_t,\bar{\mathbf m}_t$ & Feedback-best candidate configuration and its diagnostic record \\
$L_t,Q_t,e_j$ & Ordered addresses, allow-listed pair menu, and deterministic one-field edit at address $j$ \\
$\mathcal{A},a_t$ & Diagnostic-to-task association procedure and returned task address \\
$\mathcal{T}_{\mathrm{edit}}$ & Revision-request operator conditioned on recorded feedback \\
$\oplus$ & Application of a revision to a model-design record \\
$\operatorname{Select}_{\mathrm{val}}$ & Selection rule over eligible validation runs \\
$(\mathbf{h}_{\mathrm{best}},\mathbf{c}_{\mathrm{best}})$ & Validation-selected design record and executable candidate \\
\bottomrule
\end{tabularx}
\end{table*}

\FloatBarrier
\section{Detailed Case Studies of Auditable Model-Revision Trajectories}
\label{app:case_study_detail}

We reconstruct two model-revision trajectories on morphology-based perturbation-response tasks. Each record preserves execution errors, rule checks, validation outcomes, rejected candidates, and retained candidates under a fixed evaluation protocol. Phase~1--3 follow the pipeline logs: diagnosis and task setup, initial candidate generation, and subsequent review. HRT, LCA, and PDR operate throughout these phases.

On BBBC036, CellScientist retained DD-MIA (Dose-Disentangled Mechanism-Intensity Attention) after evaluating representation and fusion variants. Its exported implementation uses Morgan fingerprints, virtual-token cross-attention, fold-label adversarial regularization, and hidden-space chemical--dose gating. The architecture name records the design proposal; the executable operations are specified in Appendix~\ref{app:frozen_architecture_spec}.

On BBBC047, modulation and VAE-style candidates showed both gains and regressions. The trajectory recorded possible underuse of the supplied control-reference profile as a working diagnostic. CellScientist subsequently retained Ref-RGAF, which adds a deterministic control-reference encoding to the decoder. The record links this diagnostic to a local revision and its validation outcome.

These cases make CellScientist's selection process inspectable: the records identify attempted assumptions, failed or regressive branches, and validation-selected executable candidates. They complement the quantitative results by tracing local revisions and selection decisions under a fixed protocol.

\makeatletter
\@ifundefined{CSOverview}{%
  \newtcolorbox{CSOverview}{
    enhanced, breakable,
    colback=black!2, colframe=black!55,
    boxrule=0.9pt, arc=1.5mm,
    left=1.2mm, right=1.2mm, top=1.0mm, bottom=1.0mm,
    fonttitle=\bfseries, title=Case Study Overview
  }
}{}%
\@ifundefined{CSLoop}{%
  \newtcolorbox{CSLoop}{
    enhanced, breakable,
    colback=blue!2, colframe=blue!55,
    boxrule=0.9pt, arc=1.5mm,
    left=1.2mm, right=1.2mm, top=1.0mm, bottom=1.0mm,
    fonttitle=\bfseries, title=Closed-Loop Revision Trajectory
  }
}{}%
\@ifundefined{CSInsight}{%
  \newtcolorbox{CSInsight}{
    enhanced, breakable,
    colback=green!2, colframe=green!55,
    boxrule=0.9pt, arc=1.5mm,
    left=1.2mm, right=1.2mm, top=1.0mm, bottom=1.0mm,
    fonttitle=\bfseries, title=Key Observation
  }
}{}%
\@ifundefined{CSRisk}{%
  \newtcolorbox{CSRisk}{
    enhanced, breakable,
    colback=orange!2, colframe=orange!70!black,
    boxrule=0.9pt, arc=1.5mm,
    left=1.2mm, right=1.2mm, top=1.0mm, bottom=1.0mm,
    fonttitle=\bfseries, title=Failure Mode / Design Risk
  }
}{}%
\@ifundefined{CSLaw}{%
  \newtcolorbox{CSLaw}{
    enhanced, breakable,
    colback=purple!2, colframe=purple!60!black,
    boxrule=0.9pt, arc=1.5mm,
    left=1.2mm, right=1.2mm, top=1.0mm, bottom=1.0mm,
    fonttitle=\bfseries, title=Design Rationale Encoded as Architecture
  }
}{}%
\makeatother

\makeatletter
\@ifundefined{CSAppendixA}{%
  \newtcolorbox{CSAppendixA}{
    enhanced,
    breakable,
    colback=white,
    colframe=black!80,
    boxrule=1.2pt,
    arc=2mm,
    left=2.0mm,
    right=2.0mm,
    top=1.6mm,
    bottom=1.6mm,
    before skip=1.2em,
    after skip=1.2em,
    fonttitle=\bfseries\large,
    title={Case A: Auditable Revision Toward DD-MIA}
  }
}{}
\makeatother

\lstset{
  basicstyle=\ttfamily\small,
  columns=fullflexible,
  breaklines=true,
  frame=single,
  rulecolor=\color{black!20},
  keywordstyle=\color{blue!60!black},
  commentstyle=\color{green!40!black},
  stringstyle=\color{red!50!black},
  showstringspaces=false,
  captionpos=b
}

\clearpage
\subsection{Detailed Case Study A}
\label{app:case_a}

\begin{CSAppendixA}
{\footnotesize\itshape Trajectory including negative results, crashes, regressions, and retained candidates.}

\begin{CSOverview}
This section reconstructs the model-revision trajectory of \textbf{CellScientist} on BBBC036 under the SMILES-based split. The trajectory moves from simple feature-based modeling toward \textbf{DD-MIA}, whose learned chemical and dose gates modulate an attention update in hidden space.

\vspace{0.75em}
\noindent\textbf{What this case study illustrates.}
The trace records how CellScientist diagnosed data and batch-related issues, proposed design rationales, implemented candidates, and retained the highest-scoring eligible candidate under the validation protocol.

\vspace{0.6em}
\noindent\textbf{Trajectory evidence.}
The trajectory documents how recorded diagnostics and validation outcomes were associated with local design revisions and empirical candidate selection.
\end{CSOverview}

\begin{CSLoop}
\centering
\resizebox{\linewidth}{!}{%
\begin{tikzpicture}[
  node distance=13mm,
  every node/.style={font=\small},
  box/.style={draw,rounded corners,align=center,minimum width=32mm,minimum height=9.5mm},
  arrow/.style={-Latex,thick}
]
\node[box] (A) {Data Diagnosis\\\& Constraints};
\node[box,right=of A] (B) {Candidate\\Rationale};
\node[box,right=of B] (C) {Executable\\Synthesis};
\node[box,below=of C] (D) {Training\\\& Evaluation};
\node[box,left=of D] (E) {Revision\\\& Selection};

\draw[arrow] (A) -- (B);
\draw[arrow] (B) -- (C);
\draw[arrow] (C) -- (D);
\draw[arrow] (D) -- (E);
\draw[arrow] (E) -- (B);

\node[above=6mm of B,align=center,font=\footnotesize\bfseries] (T)
{Auditable closed-loop design record:\\diagnosis $\rightarrow$ executable candidate $\rightarrow$ validation feedback};
\end{tikzpicture}
}

\vspace{0.6em}
\noindent
\textbf{Interpretation.} The loop records a practical model-design process:
\emph{diagnose} $\rightarrow$ \emph{propose} $\rightarrow$
\emph{implement} $\rightarrow$ \emph{test} $\rightarrow$ \emph{revise}.
\end{CSLoop}

\paragraph{(A) From Representation Choice to Design Rationale}

The trajectory is non-monotonic: several candidates underperformed despite valid execution, and other candidates failed during execution. The complete record shows which components were retained, revised, or rejected at each step.

\vspace{0.5em}
Table~\ref{tab:trajectory_detail_A} summarizes representative scored candidates from the trajectory. The initial search retained a graph-based candidate, while several more complex fusion variants regressed or failed.

\begin{CSInsight}
\noindent
The trajectory preserves both validation-retained candidates and negative evidence. Several higher-capacity variants regressed, while the validation rule retained a more structured candidate. This record makes the empirical selection process auditable.
\end{CSInsight}

\paragraph{(B) Auditable Model-Revision Trajectory}

We report the diagnosis and revision phases, including failures and regressions. These records show how the validation rule selected the retained candidate.

\textbf{Phase 1: Problem Formulation and Data Diagnosis.}
Before architecture revision, the system defines the prediction target and records operational constraints. Table~\ref{tab:phase1_trace_A} summarizes the data checks that informed later candidates.

\begin{center}
\begin{minipage}{\linewidth}
\centering
\captionof{table}{\textbf{Phase 1 Execution Trace: Diagnosis and Constraint Identification.}}
\label{tab:phase1_trace_A}
\renewcommand{\arraystretch}{1.15}

\begin{adjustbox}{max width=\linewidth}
\begin{tabular}{cllp{7.0cm}}
\toprule
\textbf{Step} & \textbf{Activity}  & \textbf{Artifact} & \textbf{Design / Operational Insight} \\
\midrule
1 & Data Ingestion \& Pairing & HDF5 Database &
\textbf{Operational repair:} Resolved HDF5 sizing mismatch by enforcing type and shape alignment. Established the $f(\text{ControlRef}, \text{Chem}, \text{Dose}) \to \text{TreatedResponse}$ mapping used by later candidates. \\
2 & Statistical Diagnosis & ANOVA/PCA &
\textbf{Operational repair:} Corrected a syntax error in the statistical constraint formula.
\textbf{Data cue:} Observed a measurable plate-related signal in the feature space, motivating later checks for batch-related shortcuts. \\
3 & Dose-Response Profiling & Volcano Plot &
\textbf{Response cue:} Identified features with strong perturbation response and observed non-linear, heterogeneous dose behavior. This suggested that a purely linear response model may be insufficient. \\
4 & Technical-Factor Inspection & Assay-Variable Schematic &
\textbf{Constraint definition:} The persistence of plate effects after normalization motivated the use of batch-related regularization in later candidates. \\
\bottomrule
\end{tabular}
\end{adjustbox}
\end{minipage}
\end{center}

\begin{CSInsight}
\noindent\textbf{From diagnosis to design.}
ANOVA/PCA diagnostics associated part of the feature variation with technical factors. This recorded observation motivated batch-related regularization in later candidates.
\end{CSInsight}

\textbf{Phase 2: Generate and Execute.}
Phase 2 evaluated candidate representations and recorded setup errors and unstable training (Table~\ref{tab:phase2_trace_B}). The highest-scoring eligible candidate was recorded as a graph-based variant and exceeded the initial flat-vector candidate in validation PCC. Later revisions also changed the compound encoder; the selected DD-MIA export uses fingerprints. Phase 2 uses an exploratory baseline fit; Phase 3 aggregate gains use the benchmark reference.

\begin{center}
\begin{minipage}{\linewidth}
\centering
\captionof{table}{\textbf{Phase 2 Execution Trace: Structural Representation Trial.}}
\label{tab:phase2_trace_B}
\renewcommand{\arraystretch}{1.15}

\begin{adjustbox}{max width=\linewidth}
\begin{tabular}{clclp{6.5cm}}
\toprule
\textbf{Iter} & \textbf{Strategy} & \textbf{Outcome} & \textbf{PCC} & \textbf{Design Analysis \& Operational Event} \\
\midrule
1 & Initial Debug / Baseline$^{a}$ & Baseline & 0.3172 &
\textbf{Operational:} Pipeline stabilized through a setup-node repair.
\textbf{Baseline:} Established a reference PCC using standard flat-vector features. \\
2 & \textbf{EquiMorph-GAT} & \textbf{Retained} & \textbf{0.3660} &
\textbf{Design cue:} Explicit molecular topology encoding improved over the initial flat-vector candidate in this trace, suggesting that compound-structure representation is useful for this task. \\
3 & MD\_AdvN (Adversarial) & Regression & 0.3378 &
\textbf{Negative result:} Adversarial normalization alone obtained a lower validation PCC and was rejected by the validation rule. \\
4 & Innovation Variant & Regression & 0.3656 &
\textbf{Validation check:} A graph-based variant produced similar performance across a second implementation attempt. \\
5 & Baseline Re-test & Regression & 0.2827 &
\textbf{Regression:} The simplified baseline obtained a lower validation PCC in this trace; subsequent candidates retained the structured representation. \\
\bottomrule
\end{tabular}
\end{adjustbox}
\vspace{1pt}\par\footnotesize $^{a}$Exploratory Phase 2 baseline; Phase 3 gains use the separately refit Phase 3 reference shown in Table~\ref{tab:trajectory_detail_A}.
\end{minipage}
\end{center}

\begin{CSInsight}
\noindent\textbf{Representation cue.}
The graph-based candidate improved over simplified flat-vector baselines within this recorded trajectory, providing task-specific validation evidence for retaining compound-structure encoding.
\end{CSInsight}

\textbf{Phase 3: Review and Revise.}
Phase 3 evaluated how to combine the control-reference profile, compound representation, and dose information. Table~\ref{tab:phase3_full_trace_A} reports all iterations, including crashes and candidates rejected for lower validation performance.

\begin{center}
\begin{minipage}{\linewidth}
\centering
\captionof{table}{\textbf{Phase 3 Full Iteration Trace (Iter 1--10).}}
\label{tab:phase3_full_trace_A}
\renewcommand{\arraystretch}{1.15}

\begin{adjustbox}{max width=\linewidth}
\begin{tabular}{clclp{7.0cm}}
\toprule
\textbf{Iter} & \textbf{Strategy} & \textbf{Outcome} & \textbf{PCC} & \textbf{Design Analysis} \\
\midrule
1 & Chem-Modulated FiLM & Regression & 0.3626 &
Candidate assumption: affine modulation is sufficient. Result: validation PCC was lower than the retained graph-based model. \\
2 & Res-EquiMorph + Token & Crash & N/A &
Candidate assumption: tokenization may improve interaction modeling. Execution failed with a CUDA class-index assertion; no valid PCC was produced. \\
3 & Tokenized Adversarial & Regression & 0.3341 &
Candidate assumption: finer granularity would improve robustness. Result: the tokenized variant obtained lower validation PCC. \\
4 & DGAN (Gated-Adv) & Improved & 0.3666 &
Design pivot: adversarial regularization was restricted to the morphology branch, while chemistry was used to gate the perturbation response. \\
5 & Res-DGAN & Regression & 0.3627 &
Negative add-on: an additional residual skip produced a lower validation PCC in this run. \\
6 & Attn-DGAN (Attention) & Improved & 0.3674 &
Refinement: cross-attention provided a more structured interaction between morphology and compound features. \\
7 & Dose-Gated Attention & Regression & 0.3613 &
Negative result: directly gating attention by dose reduced validation PCC in this candidate. \\
8 & Dose-Conditioned Res & Crash & N/A &
Execution failed with a CUDA class-index assertion; no valid PCC was produced. \\
9 & Projected-Token Attn & Regression & 0.3658 &
The candidate executed successfully but remained below the retained PCC of 0.3674. \\
10 & \textbf{DD-MIA} & \textbf{Retained} & \textbf{0.3688} &
Selected design: separates compound-structure representation from dose-dependent intensity scaling using multiplicative fusion. \\
\bottomrule
\end{tabular}
\end{adjustbox}
\end{minipage}
\end{center}

\begin{CSInsight}
\noindent\textbf{Why failures matter.}
The record preserves crashes and regressions from direct dose injection into attention and more complex tokenized variants. These outcomes expose the sequence of attempted revisions and validation decisions.
\end{CSInsight}

\paragraph{(C) Validation-Selected Design: DD-MIA}

The retained \textbf{DD-MIA} candidate combines three implementation choices:

\textbf{(1) Structural compound representation.}
The selected export encodes a 2,048-bit Morgan fingerprint with an MLP, then reshapes chemical and context embeddings into 16 virtual tokens of width 32 for four-head cross-attention.

\textbf{(2) Fold-label adversarial regularization.}
Technical-variation diagnostics motivated a six-way adversarial head on the context encoding. Its labels are recorded fold IDs, with gradient-reversal strength $0.5$ and cross-entropy weight $0.1$.

\begin{CSRisk}
\noindent
The diagnostic motivated regularization against technical variation; the implemented supervision is the fold label. This identifies the training objective separately from its intended effect.
\end{CSRisk}

\textbf{(3) Chemistry--dose gating of the attention update.}
Let $h_m,h_c,h_d$ encode context, fingerprint, and dose; $a$ is the flattened context-to-chemical attention output. The selected predictor computes
\begin{equation}
u=\operatorname{LN}(h_m+a\odot\sigma(W_ch_c+b_c)\odot\sigma(W_dh_d+b_d)),
\qquad \widehat y=D([u,h_c,h_d]).
\end{equation}
The product acts on a 512-dimensional hidden update; the decoder returns the treated profile. Appendix~\ref{app:frozen_architecture_spec} gives all dimensions and losses.

\begin{CSLaw}
\noindent
The hidden gates encode a task-specific preference for multiplicative chemical--dose interaction. Retention follows the validation evidence recorded in this trajectory.
\end{CSLaw}

\paragraph{(D) Evaluation and Trajectory Audit}

The retained DD-MIA candidate achieved a five-fold validation aggregate PCC of 0.3688, corresponding to CS-model on BBBC036 in Table~\ref{tab:sota_benchmark}. The trajectory records the failed candidates, regressions, and retained components leading to this design.

\begin{center}
\centering
\resizebox{\linewidth}{!}{%
\begin{tikzpicture}[>=stealth, thick]
\draw[->] (0,0) -- (7,0) node[right]{Revision sequence (schematic)};
\draw[->] (0,0) -- (0,5) node[above]{Relative validation outcome};

\node[circle,fill=gray!40] (A) at (1,1) {\small Baseline};
\node[circle,fill=blue!40] (B) at (3,2.5) {\small EquiMorph};
\node[circle,fill=red!40] (C) at (4,2) {\small Additive Fusion};
\node[circle,fill=green!50] (D) at (5.8,4.2) {\small DD-MIA};

\draw[->] (A) -- (B);
\draw[->, dashed] (B) -- (C);
\draw[->] (C) -- (D);
\end{tikzpicture}
}

\captionof{figure}{Auditable model-revision trajectory: CellScientist follows a non-monotonic path through candidate architectures, retaining validated components while recording underperforming branches.}
\label{fig:semantic_gradient_ascent}
\end{center}

\begin{center}
\begin{minipage}{\linewidth}
\centering
\captionof{table}{\textbf{Trajectory of Candidate Performance.}}
\label{tab:trajectory_detail_A}
\renewcommand{\arraystretch}{1.2}
\resizebox{\linewidth}{!}{%
\begin{tabular}{llccc}
\toprule
\textbf{Activity Step} & \textbf{Design Rationale} & \textbf{PCC} & \textbf{$\Delta$ vs Base} & \textbf{Status} \\
\midrule
Baseline$^{a}$ & MLP (Flat Vectors) & 0.3396 & 0.00\% & Ref \\
Initial Search & EquiMorph-GAT & 0.3660 & +7.77\% & Retained candidate \\
Iter 1 & Additive FiLM & 0.3626 & +6.77\% & Regression \\
Iter 6 & Attn-DGAN & 0.3674 & +8.18\% & Improved \\
Iter 10 & DD-MIA & \textbf{0.3688} & \textbf{+8.60\%} & Retained candidate \\
\bottomrule
\end{tabular}
}
\vspace{1pt}\par\footnotesize $^{a}$Phase 3 reference fit, separately refit from the exploratory Phase 2 baseline (0.3172); all Phase 3 gains in this table use 0.3396.
\end{minipage}
\end{center}

\begin{CSInsight}
\noindent
The trajectory records both validation-retained improvements and negative evidence. DD-MIA is the candidate retained for this benchmark under the reported validation-selection rule.
\end{CSInsight}

\paragraph{(E) Summary}

The BBBC036 trajectory connects representation choices, batch-related diagnostics, and fusion revisions to recorded execution and validation outcomes. CellScientist retained DD-MIA under the fixed selection rule while preserving failed and regressive alternatives.

\end{CSAppendixA}

\makeatletter
\@ifundefined{CSAppendixB}{%
  \newtcolorbox{CSAppendixB}{
    enhanced,
    breakable,
    colback=white,
    colframe=black!80,
    boxrule=1.2pt,
    arc=2mm,
    left=2.0mm,
    right=2.0mm,
    top=1.6mm,
    bottom=1.6mm,
    before skip=1.2em,
    after skip=1.2em,
    fonttitle=\bfseries\large,
    title={Case B: Auditable Revision Toward Ref-RGAF}
  }
}{}
\makeatother

\subsection{Detailed Case Study B}
\label{app:case_b}

The introduction summarizes the BBBC047 trajectory in Figure~\ref{fig:bbbc047_trajectory}. The detailed record below links its working diagnostics, local revisions, validation outcomes, and retained candidate.

\begin{CSAppendixB}
{\footnotesize\itshape Trajectory including negative results, crashes, regressions, and retained candidates.}

\begin{CSOverview}
This section reconstructs the model-revision trajectory of \textbf{CellScientist} on BBBC047 under the SMILES-based split. The trajectory moves from standard generative baselines toward \textbf{Ref-RGAF} (Reference-Gated Affine Fusion), a candidate architecture that adds a deterministic control-reference path to the decoder.

\vspace{0.75em}
\noindent\textbf{What this case study illustrates.}
The trace records how the system evaluated modulation-based candidates, observed validation regressions in several VAE-style variants, and retained a reference-aware revision under the fixed validation protocol.

\vspace{0.6em}
\noindent\textbf{Trajectory evidence.}
The record links observed validation outcomes to subsequent local revisions and the candidate retained by the selection rule.
\end{CSOverview}

\begin{CSLoop}
\centering
\resizebox{\linewidth}{!}{%
\begin{tikzpicture}[
  node distance=13mm,
  every node/.style={font=\small},
  box/.style={draw,rounded corners,align=center,minimum width=32mm,minimum height=9.5mm},
  arrow/.style={-Latex,thick}
]
\node[box] (A) {Data Diagnosis\\\& Split Checks};
\node[box,right=of A] (B) {Candidate\\Rationale};
\node[box,right=of B] (C) {Executable\\Synthesis};
\node[box,below=of C] (D) {Training\\\& Evaluation};
\node[box,left=of D] (E) {Revision\\\& Selection};

\draw[arrow] (A) -- (B);
\draw[arrow] (B) -- (C);
\draw[arrow] (C) -- (D);
\draw[arrow] (D) -- (E);
\draw[arrow] (E) -- (B);

\node[above=6mm of B,align=center,font=\footnotesize\bfseries] (T)
{Auditable closed-loop design record:\\candidate revision under validation feedback};
\end{tikzpicture}
}
\end{CSLoop}

\paragraph{(A) From Regression to Reference-Aware Generation}

Generative candidates in the BBBC047 trajectory showed both validation gains and regressions. The run recorded possible underuse of the supplied control-reference information as a working diagnostic for subsequent candidate generation.

\vspace{0.5em}
\begin{CSInsight}
\noindent\textbf{Observed pivot.}
The retained revision injects a deterministic encoding $h$ of the supplied control-reference profile alongside the stochastic latent variable in the decoder. The observed validation result supports retaining this reference-aware design.
\end{CSInsight}

\begin{center}
\centering
\resizebox{\linewidth}{!}{%
\begin{tikzpicture}[>=stealth, thick]
\draw[->] (0,0) -- (7,0) node[right]{Revision Sequence (schematic)};
\draw[->] (0,0) -- (0,5) node[above]{Relative validation outcome};

\node[circle,fill=gray!40] (A) at (1,1) {\small Baseline MLP};
\node[circle,fill=blue!40] (B) at (3,2.0) {\small FiLM-CVAE}; 
\node[circle,fill=orange!40] (C) at (4.5,3.2) {\small MS-FiLM};
\node[circle,fill=green!50] (D) at (6.0,4.5) {\small \textbf{Ref-RGAF}};

\draw[->] (A) -- (B);
\draw[->] (B) -- (C);
\draw[->] (C) -- (D);
\end{tikzpicture}
}
\captionof{figure}{Auditable model-revision trajectory: the system moves from global modulation to multi-scale modulation and then to a reference-aware decoder.}
\end{center}

\paragraph{(B) Auditable Model-Revision Trajectory}

We report the complete trajectory, including Phase 2 instability and Phase 3 revisions, to make validation-based candidate retention inspectable.

\textbf{Phase 1: Diagnosis and Constraint Identification.}
The system identified data-quality issues and recorded preprocessing checks and split alignment before candidate evaluation (Table~\ref{tab:phase1_trace}).

\begin{center}
\begin{minipage}{\linewidth}
\centering
\captionof{table}{\textbf{Phase 1 Execution Trace: Diagnosis and Constraint Identification.}}
\label{tab:phase1_trace}
\renewcommand{\arraystretch}{1.15}
\begin{adjustbox}{max width=\linewidth}
\begin{tabular}{cllp{7.0cm}}
\toprule
\textbf{Step} & \textbf{Activity}  & \textbf{Artifact} & \textbf{Design / Operational Insight} \\
\midrule
1 & Data Ingestion & HDF5/Pandas &
\textbf{Operational repair:} Fixed a \texttt{TypeError} caused by mixed numeric and string columns in the raw CSV. \\
2 & Numerical Diagnosis & Feature Scan &
\textbf{Constraint definition:} Identified invalid-value warnings and added strict NaN checks and feature filtering to prevent training failures. \\
3 & Split Alignment & Grouped CV &
\textbf{Protocol choice:} Grouped samples by exact SMILES identity, matching the SMILES-based split used for the reported controller and trajectory scores. \\
\bottomrule
\end{tabular}
\end{adjustbox}
\end{minipage}
\end{center}

\textbf{Phase 2: Generate and Execute.}
Phase 2 recorded frequent code-generation and training failures. The baseline MLP had the highest validation PCC among eligible candidates and provided the reference for subsequent revisions (Table~\ref{tab:phase2_trace}). This Phase 2 value is an exploratory setup-trace fit; the Phase 3 aggregate gains use a separate Phase 3 reference fit.

\begin{center}
\begin{minipage}{\linewidth}
\centering
\captionof{table}{\textbf{Phase 2 Execution Trace: Establishing the Baseline.}}
\label{tab:phase2_trace}
\renewcommand{\arraystretch}{1.15}
\begin{adjustbox}{max width=\linewidth}
\begin{tabular}{clclp{6.5cm}}
\toprule
\textbf{Iter} & \textbf{Strategy} & \textbf{Outcome} & \textbf{PCC} & \textbf{Design Analysis} \\
\midrule
1 & Init & Crash & N/A & Syntax error in setup node. \\
2 & \textbf{Baseline MLP}$^{a}$ & \textbf{Retained} & \textbf{0.4118} &
\textbf{Anchor candidate:} A standard MLP using median-normalized features provided a stable reference point. \\
3 & Innovation Var. & Regression & 0.3834 &
\textbf{Negative result:} A more complex candidate obtained a lower validation PCC and was rejected by the validation rule. \\
4 & Training Loop & Repaired / Regression & 0.3707 &
\textbf{Instability:} Training-node failure was repaired, but the candidate degraded. \\
5 & Final Attempt & Repaired / Regression & 0.3468 &
\textbf{Regression:} Severe instability led the system to preserve Iteration 2 as the validation-selected anchor. \\
\bottomrule
\end{tabular}
\end{adjustbox}
\vspace{1pt}\par\footnotesize $^{a}$Exploratory Phase 2 baseline; the Phase 3 trajectory uses the separately refit reference of 0.4120 shown in Table~\ref{tab:trajectory_detail}.
\end{minipage}
\end{center}

\textbf{Phase 3: Review and Revise.}
Phase 3 evaluated modulation, gating, and reference-aware variants (Table~\ref{tab:phase3_full_trace}). The retained Ref-RGAF candidate adds a deterministic control-reference path to the generative architecture.

\begin{center}
\begin{minipage}{\linewidth}
\centering
\captionof{table}{\textbf{Phase 3 Full Iteration Trace: Revision Logic.} ``Improved'' denotes an increase from the immediately preceding candidate; ``Retained'' denotes the best validation-selected candidate in this trace.}
\label{tab:phase3_full_trace}
\renewcommand{\arraystretch}{1.15}
\begin{adjustbox}{max width=\linewidth}
\begin{tabular}{clclp{7.0cm}}
\toprule
\textbf{Iter} & \textbf{Strategy} & \textbf{Outcome} & \textbf{PCC} & \textbf{Design Analysis} \\
\midrule
1 & \textbf{FiLM-CVAE} & Improved & 0.4148 &
\textbf{Pivot:} Replaced a rigid ODE-style candidate with FiLM modulation, where drug embedding modulates the latent state. \\
2 & Res-FiLM & Regression & 0.3416 &
\textbf{Negative result:} The residual VAE obtained a lower validation PCC; the trace used possible underuse of control-reference information as a working diagnostic for later revisions. \\
3 & Dose-Gated & Improved & 0.4095 &
Candidate assumption: simple dose gating may be sufficient. Result: the candidate obtained a lower validation PCC than the retained anchor. \\
4 & TransMod & Improved & 0.4097 &
The transformer-style fusion candidate obtained a lower validation PCC than the retained anchor. \\
5 & \textbf{MS-FiLM} & Improved & \textbf{0.4237} &
\textbf{Design cue:} The multi-layer modulation candidate exceeded the bottleneck-only candidate in validation PCC. \\
6 & Gated-Attn & Improved & 0.4265 &
Multiplicative gating improved validation PCC relative to the preceding candidates in this trace. \\
7 & RGAF-FiLM & Improved & 0.4271 &
Added affine fusion and a residual hypernetwork. \\
8 & \textbf{Ref-RGAF} & \textbf{Retained} & \textbf{0.4368} &
\textbf{Key revision:} Concatenated deterministic control-reference encoding $h$ with latent $z$; this candidate obtained the highest validation PCC in the trace. \\
9 & State-Adaptive & Regression & 0.4249 &
The state-adaptive fusion candidate obtained a lower validation PCC than Ref-RGAF. \\
10 & LoRA-FiLM & Regression & 0.4163 &
The LoRA-style latent-rotation candidate obtained a lower validation PCC than Ref-RGAF. \\
\bottomrule
\end{tabular}
\end{adjustbox}
\end{minipage}
\end{center}

\begin{CSRisk}
\noindent\textbf{Working diagnostic.}
The residual VAE candidate used $y=x+\text{Dec}(z)$ and obtained a lower validation PCC. The run record proposed possible underuse of the supplied control-reference information as a working diagnostic and used it to motivate reference injection.
\end{CSRisk}

\paragraph{(C) Validation-Selected Design: Reference-Gated Affine Fusion}

The retained \textbf{Ref-RGAF} candidate combines three implementation choices:

\textbf{(1) Control-reference conditioning.}
Ref-RGAF adds a deterministic reference path:
\begin{equation}
v_0 = [z_{\text{stochastic}}, h_{\text{deterministic}}].
\end{equation}
The joint state $v_0$ is affinely modulated before decoding, preserving a deterministic encoding of the supplied plate-matched DMSO control profile alongside the sampled latent state.

\textbf{(2) Affine dose modulation.}
The drug effect is represented as a dose-conditioned affine transformation:
\begin{equation}
\text{DrugFeat} = (E_{\text{chem}} \odot \sigma(\text{Scale}(d))) + \text{Shift}(d).
\end{equation}
Layer normalization and SiLU transform this intermediate feature before a hypernetwork supplies scale and shift vectors for three decoder stages.

\textbf{(3) Multi-scale modulation.}
Drug modulation is applied at multiple decoder layers, allowing perturbation effects to enter at several representation levels.

\begin{CSLaw}
\noindent\textbf{Design rationale.}
The retained architecture reflects a task-specific modeling preference for preserving deterministic control-reference information while combining it with compound and dose features.
\end{CSLaw}

\paragraph{(D) Validation Performance and Response-Sensitive Metrics}

The retained Ref-RGAF candidate achieved a five-fold validation aggregate PCC of 0.4368 (Table~\ref{tab:trajectory_detail}), corresponding to CS-model on BBBC047 in Table~\ref{tab:sota_benchmark}. This reference-aware revision was retained by the validation-selection rule.

\begin{center}
\begin{minipage}{\linewidth}
\centering
\captionof{table}{\textbf{Trajectory of Candidate Performance (Phase 3).}}
\label{tab:trajectory_detail}
\renewcommand{\arraystretch}{1.2}
\resizebox{\linewidth}{!}{%
\begin{tabular}{llcccc}
\toprule
\textbf{Iter} & \textbf{Architecture Strategy} & \textbf{PCC} & \textbf{$\Delta$ vs Base} & \textbf{RF-RMSE $\downarrow$} & \textbf{Status} \\
\midrule
Ref$^{a}$ & Baseline MLP & 0.4120 & 0.00\% & 5.13 & Reference \\
1 & FiLM-CVAE & 0.4148 & +0.68\% & 5.07 & Early gain \\
5 & MS-FiLM & 0.4237 & +2.84\% & 5.01 & Improved \\
\textbf{8} & \textbf{Ref-RGAF} & \textbf{0.4368} & \textbf{+6.02\%} & \textbf{4.96} & \textbf{Retained} \\
\bottomrule
\end{tabular}
}
\vspace{1pt}\par\footnotesize $^{a}$Phase 3 reference fit, separately refit from the exploratory Phase 2 baseline (0.4118); all Phase 3 gains in this table use 0.4120.
\end{minipage}
\end{center}

\begin{CSInsight}
\noindent\textbf{Interpreting RF-RMSE.}
Response-feature RMSE measures prediction error on the most perturbation-responsive morphology features, complementing global PCC when evaluating the retained candidate.
\end{CSInsight}

\paragraph{(E) Summary}

On BBBC047, CellScientist used a working diagnostic about control-reference use to propose the Ref-RGAF design. The trajectory preserves valid but underperforming candidates and links the reference-aware revision to execution results and validation-based selection.
\end{CSAppendixB}

\section{Experimental Details}
\label{app:exp_details}

This appendix describes the datasets, baselines, metrics, and training and evaluation protocols used in Section~\ref{sec:exp_setup}.

\subsection{Benchmark Tasks and Datasets}
\label{app:benchmarks}

\subsubsection{Cell Painting Morphology Prediction}
\label{app:exp_morph}

We evaluate model discovery on BBBC036, BBBC047, and CPG0016, and transfer two frozen designs to LKCP Batch 2. Prediction targets are CellProfiler features~\citep{stirling2021cellprofiler} describing cellular size, shape, texture, and fluorescence intensity. Each task uses an assay-context reference derived from vehicle controls according to its preprocessing protocol.

BBBC036 and BBBC047 derive from the Cell Painting small-molecule collection of Bray et al.~\citep{bray2017dataset}. CPG0016 is part of the JUMP Cell Painting resource~\citep{chandrasekaran2023jump,arevalo2024evaluating}. All three use U2OS cells and span progressively larger sets of chemical perturbations. Table~\ref{tab:pheno_datasets} reports source-profile counts for the processed inputs, distinguishing raw profiles from treated observations used for response prediction. Raw profiles include vehicle controls; treated profiles contain only retained non-DMSO observations. Exact non-control SMILES are counted within each benchmark, and raw feature counts precede split-specific filtering.

\begin{table*}[t]
\centering
\caption{\textbf{Source-level statistics of the processed Cell Painting inputs used in our experiments.} Raw profiles include vehicle controls. Treated profiles are non-DMSO observations retained for response prediction. Exact non-control SMILES are counted within each benchmark, and raw Cell Painting feature counts precede split-specific feature filtering.}
\label{tab:pheno_datasets}
{\scriptsize
\setlength{\tabcolsep}{3pt}
\renewcommand{\arraystretch}{1.05}
\begin{tabular}{lcrrrrrr}
\toprule
\textbf{Dataset} & \textbf{Cell line} & \shortstack{\textbf{Raw}\\\textbf{profiles}} & \shortstack{\textbf{DMSO}\\\textbf{controls}} & \shortstack{\textbf{Treated}\\\textbf{profiles}} & \shortstack{\textbf{Exact non-control}\\\textbf{SMILES}} & \shortstack{\textbf{Plates}\\\textbf{(all / treated)}} & \shortstack{\textbf{Raw CP}\\\textbf{features}} \\
\midrule
BBBC036 & U2OS & 18,540 & 3,528 & 15,012 & 1,916 & 55 / 47 & 602 \\
BBBC047 & U2OS & 59,275 & 16,302 & 42,973 & 11,415 & 241 / 229 & 894 \\
CPG0016 & U2OS & 235,307 & 32,837 & 202,470 & 82,282 & 407 / 392 & 737 \\
\bottomrule
\end{tabular}
}
\end{table*}

The discovery benchmarks use two group-based holdout settings. The SMILES-based setting groups records by exact recorded SMILES, and the plate-based setting groups records by experimental plate. Complete groups are assigned to a common fold.

LKCP Batch 2 comes from the LINCS Cell Painting release~\citep{natoli2021lincs,way2022morphology}
in the Cell Painting Gallery~\citep{weisbart2024gallery}. We use a processed
subset of the December 5, 2017 batch containing 64 compounds, 1,728 conditions,
and 1,671 morphology features. Compound-level hashing assigns 33 compounds
(891 conditions) to training, 15 (405 conditions) to validation, and 16
(432 conditions) to testing. Inputs are a same-plate median vehicle-control
profile, a 2,048-bit radius-2 Morgan fingerprint, and log dose. The treated-scale
target is reconstructed by adding the condition-median treated-minus-control
profile to the condition-median control profile.

\subsubsection{LINCS2020 Transcriptomic Perturbation Prediction}
\label{app:exp_gene}

For transcriptomic perturbation prediction, we use gene-expression profiles from the next-generation Connectivity Map (CMap) project, which measures large-scale transcriptional responses with the L1000 platform~\citep{subramanian2017high}. Raw and processed profiles are obtained through the LINCS data portal~\citep{Keenan2020}, following recommended access and preprocessing protocols~\citep{xie2022getting}. We construct two benchmark variants to evaluate different generalization regimes: a controlled \textbf{Seven Cell Lines} subset and a large-scale \textbf{Full Data} benchmark.

The \textbf{Seven Cell Lines} subset contains 355 compounds profiled across seven frequently assayed cell lines: A375, HELA, PC3, MCF7, HT29, YAPC, and HA1E. This dense subspace provides high compound--cell coverage and enables a chemical-blind split, where scored compounds are unseen in the corresponding training folds. It therefore measures generalization to held-out compound identities. The \textbf{Full Data} benchmark contains 8,316 compounds across 164 cell lines, yielding 78,569 compound--cell expression profiles. This setting is substantially sparser and more heterogeneous and measures performance under long-tail cell-line coverage and diverse perturbation contexts. Dataset statistics are summarized in Table~\ref{tab:lincs_datasets}.

\begin{table}[h]
\centering
\caption{\textbf{Statistics of LINCS L1000 benchmarks.}}
\label{tab:lincs_datasets}
\begin{tabular}{lcccc}
\toprule
\textbf{Dataset} & \textbf{\# Compounds} & \textbf{\# Cell Lines} & \textbf{\# Samples} & \textbf{Characteristics} \\
\midrule
Seven Cell Lines & 355 & 7 & $\sim$2,485 & Dense; chemical-blind \\
Full Data & 8,316 & 164 & 78,569 & Sparse; long-tail \\
\bottomrule
\end{tabular}
\end{table}

\subsubsection{Single-cell Perturbation Prediction}
\label{app:exp_singlecell}

We further evaluate the CellScientist workflow on public single-cell benchmarks covering genetic perturbation and multimodal perturbation responses. These datasets complement morphology and L1000 evaluations with measurements at single-cell resolution and response-sensitive prediction targets.

\textbf{Norman}~\citep{norman2019exploring} is a pooled CRISPRa scRNA-seq dataset designed to study the effects of single and combinatorial gene activation. It is commonly used to evaluate transcriptional-response prediction for genetic interventions, including combinations. \textbf{Papalexi}~\citep{papalexi2021characterizing} is a CRISPR--Cas9 ECCITE-seq dataset that jointly profiles RNA and surface-protein responses after genetic perturbation, enabling evaluation on both transcriptomic and antibody-derived-tag readouts. Table~\ref{tab:singlecell_datasets} summarizes the processed benchmark statistics used in the corresponding evaluations.

\begin{table*}[t]
\centering
\caption{\textbf{Statistics of single-cell benchmarks used in our evaluations.} Readout dimensions are reported where the evaluation uses a fixed feature matrix; Papalexi RNA and protein are evaluated as separate readouts.}
\label{tab:singlecell_datasets}
{\footnotesize
\setlength{\tabcolsep}{5pt}
\renewcommand{\arraystretch}{1.08}
\begin{tabularx}{\linewidth}{@{}l>{\raggedright\arraybackslash}p{0.28\linewidth}rX@{}}
\toprule
\textbf{Benchmark} & \textbf{Assay / task} & \textbf{Cells} & \textbf{Readout and condition statistics} \\
\midrule
Norman & CRISPRa scRNA-seq & 91,205 & 5,045 RNA genes; 284 conditions (152 single-gene, 131 two-gene, and 1 control) \\
Papalexi & CRISPR--Cas9 ECCITE-seq & 20,729 & RNA and surface-protein ADT readouts; 25 targeted genes plus non-targeting control and 98 perturbation labels \\
\bottomrule
\end{tabularx}
}
\end{table*}

We report task-specific workflow results on Norman, Papalexi RNA, and Papalexi Protein in Appendix~\ref{app:res_trans}. Each readout is evaluated under its dataset-specific protocol.

\subsection{Baselines}
\label{app:baselines}

\subsubsection{Morphology Baselines}
\label{app:morph_baselines}

For morphology prediction, we organize baselines into three comparison groups used consistently in the result tables.

\textbf{Existing reference methods} include strong feature-space predictors:
\begin{itemize}
    \item \textbf{TabR}~\citep{gorishniy2024tabr}: A tabular prediction model with a retrieval component that uses nearest-neighbor information at inference time.
    \item \textbf{RealMLP}~\citep{holzmuller2024better}: A carefully tuned MLP baseline designed to provide strong and stable performance on tabular benchmarks.
    \item \textbf{RF-TD}~\citep{holzmuller2024better}: A Random Forest baseline with tuned default settings for competitive tabular prediction.
\end{itemize}

\textbf{Search-based baselines} provide generic model-selection reference points under the corresponding benchmark split, target, and metric definitions:
\begin{itemize}
    \item \textbf{FLAML}~\citep{wang2021flaml}: An AutoML framework that searches over model and hyperparameter configurations under a configured budget.
    \item \textbf{Random Search}~\citep{gijsbers2024amlb}: A generic search baseline that samples candidate configurations independently.
\end{itemize}

\textbf{Agent-designed baselines} provide a comparison with an existing automated modeling workflow:
\begin{itemize}
    \item \textbf{CellForge}~\citep{tang2025cellforge}: A multi-agent automated modeling framework for proposing, training, and selecting predictive models.
\end{itemize}

The LKCP comparison uses RealMLP, RF-SKL, early-fusion MLP, and control-only MLP.
RF-SKL is a scikit-learn random forest~\citep{breiman2001random,pedregosa2011scikit},
whereas the discovery benchmarks use the tuned-default RF-TD variant.
RealMLP, RF-SKL, and early-fusion MLP concatenate context, fingerprint, and dose.
Control-only MLP uses the same MLP with chemical and dose inputs fixed to zero.
The two transferred CS-models preserve the DD-MIA and Ref-RGAF architectures
specified in Appendix~\ref{app:frozen_architecture_spec}, adapting their
morphology input and output dimensions to LKCP.

\subsubsection{Transcriptomic and Single-cell Baselines}
\label{app:trans_singlecell_baselines}

For LINCS2020 transcriptomic perturbation prediction, we compare against representative domain-specific baselines:
\begin{itemize}
    \item \textbf{DLEPS}~\citep{zhu2021dleps}: A deep-learning predictor of chemically induced transcriptional changes for drug screening.
    \item \textbf{DeepCE}~\citep{pham2021deepce}: A graph neural network framework that combines chemical substructure representations with cell-line features.
    \item \textbf{CIGER}~\citep{pham2022ciger}: An attention-based model that predicts gene-expression rankings from chemical structures.
    \item \textbf{MultiDCP}~\citep{wu2022multidcp}: A multi-task model of dose-dependent, context-specific gene-expression and cell-viability responses.
    \item \textbf{TranSiGen}~\citep{tong2024transigen}: A VAE-based generative model that learns latent representations for denoising and reconstructing perturbation profiles.
\end{itemize}

For the workflow-instantiation results on Norman and Papalexi, we compare against \textbf{CellForge}~\citep{tang2025cellforge} under the dataset-specific evaluation procedures.

\subsection{Evaluation Metrics}
\label{app:metrics}

\subsubsection{Morphology Metrics}
\label{app:morph_metrics}

For morphology prediction, we evaluate global profile fidelity and perturbation-sensitive feature recovery. Given prediction and target matrices $\widehat{\mathbf{Y}}$ and $\mathbf{Y}$, global PCC is $\operatorname{corr}(\operatorname{vec}(\widehat{\mathbf{Y}}),\operatorname{vec}(\mathbf{Y}))$. MSE measures numerical reconstruction error, PCC measures profile association, and $R^2$ measures explained variation relative to feature means. The discovery benchmarks summarize validation scores across five folds; the controller audit reports the validation-selected candidate on its reporting fold. LKCP uses global PCC, entry-wise MSE, and mean feature-wise $R^2$ on the fixed external test partition, summarized as mean$\pm$sample SD over five neural-model refits and a point estimate for the single RF-SKL fit. Negative feature-wise $R^2$ indicates greater squared error than predicting that feature's test-set mean; PCC is insensitive to affine scale and offset. Appendix~\ref{app:run_level_uncertainty} reports seed-matched differences from RealMLP.

To further assess whether a model captures the most responsive morphology changes, we compute response-feature metrics on the top-$K$ perturbed features. For each sample $i$, let $\mathbf{y}_{i,\mathrm{gt}}$ be the ground-truth response vector and $\boldsymbol{\mu}_{\mathrm{ctrl}}$ the mean control profile. We select responsive features using only ground-truth deviation from control:
\[
\mathcal{F}_i^{(K)}
=
\operatorname{topk}_{j}
\left(
\left|
\mathbf{y}_{i,\mathrm{gt}}^{(j)}
-
\boldsymbol{\mu}_{\mathrm{ctrl}}^{(j)}
\right|
\right),
\qquad
K\in\{20,50\}.
\]
We then compute $\mathrm{RMSE}_{\mathrm{RF}\text{-}K}$ and $\mathrm{PCC}_{\mathrm{RF}\text{-}K}$ within $\mathcal{F}_i^{(K)}$, where RF denotes response feature. The feature subset is determined from ground truth and control profiles. These metrics complement the global PCC used for discovery and controller retention.

\subsubsection{Transcriptomic and Single-cell Metrics}
\label{app:trans_singlecell_metrics}

For LINCS2020 transcriptomic prediction, RMSE and PCC measure global expression-profile fidelity. Positive and Negative P@100 measure recovery of the most strongly upregulated and downregulated genes, respectively. These metrics complement whole-profile accuracy with the recovery of strongly responding genes.

For single-cell perturbation prediction, we report PCC and $\mathrm{PCC}_{\mathrm{DE}}$. PCC measures global agreement between predicted and observed response profiles, while $\mathrm{PCC}_{\mathrm{DE}}$ evaluates the designated differentially expressed subset under each dataset-specific protocol. Together, these metrics separate overall profile reconstruction from recovery of key response signals.

\subsection{Implementation Details}
\label{app:implementation_details}

\subsubsection{Frozen Predictor Implementations}
\label{app:frozen_architecture_spec}

The frozen predictors use the following architectures. Let $x\in\mathbb R^p$ be the supplied control profile, $c\in\{0,1\}^{2048}$ the radius-2 Morgan fingerprint, and $d=\log_{10}(\mathrm{dose}+10^{-6})$. External LKCP refits use $p=1671$ and millimolar dose. The architectures are retained while input/output dimensions adapt to the cohort.

\paragraph{DD-MIA.}
The context and chemical encoders map $p\to512$ and $2048\to512$, respectively, each with BatchNorm, ReLU, and two residual blocks. A residual block computes $u+\operatorname{Dropout}_{0.2}(\operatorname{ReLU}(\operatorname{BN}(Wu+b)))$. The dose MLP is $1\to64\to64$ with an intermediate ReLU. The context and chemical 512-vectors are reshaped into $16\times32$ tokens; four-head attention uses context queries and chemical keys/values. Its flattened output $a\in\mathbb R^{512}$ is multiplied by separate chemical and dose sigmoid gates, then added to the context embedding and layer-normalized. The predictor consumes the concatenated $[u,h_c,h_d]\in\mathbb R^{1088}$ through $1088\to512\to256\to p$, with BatchNorm/ReLU/dropout $0.2$ after the first layer and ReLU after the second. It directly outputs the treated profile.

A fold classifier applies gradient reversal of strength $0.5$ to $h_m$, followed by $512\to128\to6$ with ReLU/dropout $0.2$. The training objective is
\[
\mathcal L_{\mathrm{DD}}=\operatorname{MSE}(\widehat y,y)
+0.1\operatorname{CE}(C(\operatorname{GRL}_{0.5}(h_m)),f-1),
\]
where $f$ is the recorded fold ID. The classifier minimizes cross-entropy, while gradient reversal opposes its gradient into the context encoder. The six-way head is retained on LKCP, whose training data use three fold labels.

\paragraph{Ref-RGAF.}
The context encoder is $p\to1024\to256$, with BatchNorm and SiLU after each layer, giving deterministic state $h$. Two $256\to64$ heads yield $\mu$ and $\log\sigma^2$; $z=\mu+\sigma\odot\epsilon$, with $\epsilon\sim\mathcal N(0,I)$. A $2048\to256$ chemical encoder uses LayerNorm and SiLU. Separate $1\to64\to256$ dose MLPs with SiLU produce scale $S(d)$ and shift $B(d)$:
\[
q=\operatorname{SiLU}(\operatorname{LN}(E_c(c)\odot\sigma(S(d))+B(d))).
\]
A hypernetwork maps $256\to512$, applies SiLU and two residual blocks, then maps $512\to3200$. Each block contains two linear--LayerNorm--SiLU--dropout($0.1$) layers with a residual addition. The output supplies scale/shift pairs at widths 320, 256, and 1024. The initial state $[z,h]$ has width 320. Decoder dimensions are $320\to256\to1024\to p$, with BatchNorm/SiLU at the two hidden layers. Modulation $v\mapsto v\odot(1+\gamma_j)+\beta_j$ occurs before decoding and after each hidden activation. The final layer is linear, and the objective is
\[
\mathcal L_{\mathrm{Ref}}=\operatorname{MSE}(\widehat y,y)+0.001\mathcal L_{\mathrm{KL}},
\qquad\mathcal L_{\mathrm{KL}}=-\tfrac12\operatorname{mean}(1+\log\sigma^2-\mu^2-\sigma^2).
\]
The KL mean covers batch and latent entries. Sampling remains active during validation and testing. The fixed refit seed determines evaluation random states: seed+$100000$+epoch during validation, seed+$300000$ for saved-checkpoint validation, and seed+$500000$ for test evaluation.

\subsubsection{Implementation and Execution Configuration}
\label{app:implementation_config}

Table~\ref{tab:system_specs} reports the environments used for the high-freedom workflow and the controlled audit. Methods compared within each experiment use the same hardware and software environment. Table~\ref{tab:detailed_hyperparams} separates open-workflow settings from those used to compare controllers within the fixed candidate language.

For LKCP refitting, DD-MIA uses training-standardized context, Adam at $10^{-4}$,
batch size 32, and its adversarial objective. Ref-RGAF uses unscaled context,
Adam at $10^{-3}$, batch size 64, and MSE plus $0.001$ times KL loss.
Both use at most 500 epochs and validation-MSE early stopping with patience 50;
Ref-RGAF retains latent sampling with fixed seed-derived random states at evaluation.
DD-MIA's validation loss averages mini-batch MSEs equally; Ref-RGAF uses entry-wise MSE.
RealMLP uses the PyTabKit tuned-default configuration with at most 500 epochs,
batch size 32, and early stopping. RF-SKL uses 200 trees. Context and response
scalers for the tabular predictors are fitted on training data, and predictions
are returned to the target scale. Neural refits use seeds 1729--1733; RF-SKL
uses one fixed seed. Model designs remain fixed throughout external refitting.

\begin{table}[h]
\centering
\caption{\textbf{Computational infrastructure and software environment.}}
\label{tab:system_specs}
\small
\resizebox{\linewidth}{!}{%
\begin{tabular}{ll}
\toprule
\textbf{Component} & \textbf{Specification} \\
\midrule
High-freedom workflow GPU & NVIDIA RTX 5880 Ada Generation (48GB VRAM) \\
High-freedom workflow software & Python 3.11.14; PyTorch 2.0.1+cu118 \\
Controlled-audit GPU & NVIDIA H100 (80GB HBM3) \\
Controlled-audit software & Python 3.10.20; PyTorch 2.3.1+cu121 \\
Operating system & Linux \\
\bottomrule
\end{tabular}
}
\end{table}

\begin{table*}[t]
\centering
\caption{\textbf{CellScientist controller and execution settings.}}
\label{tab:detailed_hyperparams}
\footnotesize
\renewcommand{\arraystretch}{1.02}
\setlength{\tabcolsep}{4pt}
\begin{tabularx}{\textwidth}{@{}>{\raggedright\arraybackslash}p{0.25\textwidth}>{\raggedright\arraybackslash}p{0.18\textwidth}X@{}}
\toprule
\textbf{Configuration} & \textbf{Value} & \textbf{Role in the workflow} \\
\midrule
Open-workflow backbone & Gemini 3 Pro & Default backbone for high-freedom model exploration \\
Open-workflow context length & 40,000 tokens & Maintained interaction context limit \\
Open-workflow temperatures & 0.5 analysis; 0.7 proposal & Sampling settings for analysis and generative revision \\
Open-workflow prompt views & 4 & Robust scaling, PCA, feature-network, and enrichment views \\
Open-workflow iteration budget & 10 revisions & Maximum number of high-freedom revision steps \\
Open-workflow repair budget & 3 rounds & Configured debugging budget for runtime or interface errors \\
Controlled-audit backbones & Gemini 3 Pro; Qwen2.5-0.5B-Instruct & Manuscript-default and open-weight reproduction \\
Controlled-audit decoding & Temperature 0 & Fixed within every matched controller comparison \\
Controlled-audit candidate language & 768 executable candidates & Shared predictor configurations available to every policy \\
Controlled-audit fitting & Seeded randomized low-rank fitting & Identical fitting seed for every policy within each paired run \\
Controlled-audit budgets & $B\in\{1,3,5,10\}$ & Trained and evaluated candidates, including $h_0$ \\
Validation selection metric & Global PCC & Metric used by $\operatorname{Select}_{\mathrm{val}}$ in the controlled audit \\
Controlled-audit retention view & Mean feedback/selection PCC & Identical candidate-retention rule for every matched policy \\
\bottomrule
\end{tabularx}
\end{table*}

\subsubsection{Protocol Controls and Routing Details}
\label{app:protocol_controls}

This section specifies evaluation semantics and feedback routing for the high-freedom workflow. Within each benchmark, methods use the reported split, target, and metric definitions. Fixed predictors follow their model protocols, search baselines use configured search procedures, and CellForge follows its agent workflow. CellScientist uses 10 revision steps with up to 3 local implementation repairs per step. Candidates are eligible for selection only after passing the configured checks and completing scoring; rejected and failed candidates remain in the run history.

PDR associates execution and validation diagnostics with proposed component edits. The open-code prompt contains baseline/candidate metric JSON, mutable and read-only notebook context, history, and a task-graph summary. Deterministic text flags improved PCC with worse MSE as a scale/bias discrepancy, improved MSE with worse PCC as conservative fitting, and worsening top-response-feature error as a weighting issue. The first two flags are mutually exclusive; the feature-error flag is independent. The LLM returns a focus, task names, and cell-indexed edits. Active tasks are resolved by returned-name matching, then focus-label matching, then root children; matching is case-insensitive and allows substrings. Initial edits are applied only to registered mutable cells. Graph dependencies supply context and attribution rather than an additional edit-validity test. Runtime repair uses error context and a bounded retry prompt. Table~\ref{tab:pdr_routing_examples} summarizes diagnostic-to-edit mappings; Listing~\ref{lst:pdr_json_example} summarizes the early diagnostic and selected design in one recorded trajectory.

\begin{table}[h]
\centering
\caption{
\textbf{Representative PDR mappings.}
PDR maps diagnostic signals to proposed local edit targets; LCA checks the resulting edits against protected evaluation semantics.
}
\label{tab:pdr_routing_examples}
\renewcommand{\arraystretch}{1.08}
\setlength{\tabcolsep}{4pt}
\begin{tabularx}{\linewidth}{
  >{\raggedright\arraybackslash}p{0.23\linewidth}
  >{\raggedright\arraybackslash}X
  >{\raggedright\arraybackslash}X}
\toprule
\textbf{Diagnostic} 
& \textbf{Editable region} 
& \textbf{Protected semantics} \\
\midrule
Interface failure 
& Data adapter or implementation interface 
& Split, target, and metric definitions \\
Metric-reporting failure 
& Evaluation adapter 
& Metric formulas and fold protocol \\
Validation collapse 
& Fusion, decoder, objective, or training node 
& Dataset split, target definition, and output space \\
Weak perturbation sensitivity 
& Perturbation encoding, dose encoding, or fusion edge 
& Evaluation code and response-feature definition \\
Suspected underuse of control reference 
& Encoder state, decoder input, or fusion component 
& Preprocessing, split, target, and metrics \\
Batch-shortcut risk 
& Representation or training-regularization component 
& Configured split and evaluation rules \\
\bottomrule
\end{tabularx}
\end{table}

\FloatBarrier
\begin{lstlisting}[caption={Structured summary of the BBBC047 trajectory: the iteration-2 diagnostic and the iteration-8 Ref-RGAF candidate.},label={lst:pdr_json_example}]
{
  "dataset": "BBBC047",
  "iteration": 2,
  "diagnostic": {
    "type": "suspected_control_reference_underuse",
    "candidate": "Res-FiLM",
    "execution_status": "valid",
    "rule_status": "passed",
    "selection_metric": "PCC",
    "candidate_score": 0.3416,
    "best_previous_score": 0.4148,
    "verdict": "valid_but_regressed",
    "summary": "Validation PCC dropped after a residual VAE-style update; control-reference use was recorded as a working diagnostic."
  },
  "routed_address": {
    "address_type": "proposed_edit_targets",
    "editable_components": [
      "model_architecture",
      "encoder_state_representation",
      "decoder_input",
      "fusion_module"
    ],
    "routing_reason": "valid execution with degraded validation performance and a working diagnostic about control-reference use"
  },
  "allowed_edit": {
    "operation": "test_reference_aware_decoder_path",
    "protected_components": [
      "dataset_split",
      "target_definition",
      "feature_preprocessing",
      "validation_protocol",
      "evaluation_metrics",
      "fold_protocol"
    ]
  },
  "revised_candidate": {
    "iteration": 8,
    "name": "Ref-RGAF",
    "design_change": "concatenate deterministic control-reference encoding h with stochastic latent code z before decoding",
    "intended_effect": "retain supplied control-reference information while allowing compound-dose modulation"
  },
  "lca_check": [
    "interface_check",
    "shape_check",
    "configured_rule_check",
    "metric_completeness"
  ]
}
\end{lstlisting}
\FloatBarrier

\subsubsection{Shared Candidate Language and Policy Access}
\label{app:controlled_language}

\paragraph{Domains and one-edit neighborhoods.}
Each candidate is a tuple $(u,v,r_x,r_y,\alpha)$ specifying the input blocks, response target, input and output PCA dimensions, and regularization strength of the common low-rank predictor. Let $c$ denote the supplied control-reference profile, $p$ the hashed SMILES descriptor, and $a$ log dose. Direct prediction fits $y$; response prediction fits $y-c$ and adds $c$ back at inference. The full Cartesian product contains $4\times2\times4\times4\times6=768$ distinct candidate IDs. The common initial candidate is $([c,a],y-c,32,32,10)$. Requested PCA dimensions are capped by the available sample and feature dimensions during fitting.

\begin{table}[!ht]
\centering
\caption{\textbf{The shared 768-candidate language.} The listed order is also the fixed preference order used by CellScientist and Random routing to instantiate an edit within an already selected address. Alternatives count one-field changes from an incumbent before previously evaluated IDs are removed.}
\label{tab:controlled_candidate_language}
\small
\setlength{\tabcolsep}{5pt}
\renewcommand{\arraystretch}{1.16}
\begin{tabularx}{\linewidth}{@{}lXcc@{}}
\toprule
\textbf{Address} & \textbf{Allowed values, in edit-preference order} & \textbf{Choices} & \textbf{Alternatives} \\
\midrule
Conditioning & $[c,p,a],\ [c,p],\ [c,a],\ c$ & 4 & 3 \\
Response target & $y-c,\ y$ & 2 & 1 \\
Input representation & $64,\ 128,\ 32,\ 16$ & 4 & 3 \\
Output decoder & $64,\ 128,\ 32,\ 16$ & 4 & 3 \\
Regularization & $1,\ 0.1,\ 0.01,\ 10,\ 100,\ 1000$ & 6 & 5 \\
\bottomrule
\end{tabularx}
\end{table}

An atomic transition changes exactly one tuple coordinate. Every state therefore has $3+1+3+3+5=15$ one-edit neighbors before history filtering. These are local neighborhood counts, not disjoint subsets of the global language. For incumbent $h$ and previously evaluated IDs $S$, the available address set is
\[
\mathcal{A}_{\mathrm{avail}}(h,S)
=\{j:\mathcal{N}_j(h)\setminus S\ne\varnothing\},
\]
where $\mathcal{N}_j(h)$ contains the one-field alternatives at address $j$. Random routing samples uniformly from this set. CellScientist ranks addresses within it using diagnostics and history, then exposes the first two ranked addresses. Both instantiate one unseen candidate at an exposed address using the same preference order in Table~\ref{tab:controlled_candidate_language}, with candidate ID as the tie-breaker. For identical $(h,S,j)$, the instantiated edit is identical under both policies. Allow-list and pairing checks constrain proposals before evaluation; the runner records transition admissibility.

\paragraph{What each policy receives.}
The shared object is the candidate universe and its evaluator. The per-step candidate menu is produced by the search policy, as specified in Table~\ref{tab:controlled_policy_access}. CellScientist's controlled implementation routes among existing configurations; candidate generation in the open-code workflow is evaluated separately.

\begin{table}[!ht]
\centering
\caption{\textbf{Policy-specific access to the common candidate universe.} Every revising policy uses the same $h_0$, fitting procedure, feedback/selection/test partitions, and total candidate budget. Unseen means not evaluated earlier in that trajectory.}
\label{tab:controlled_policy_access}
\small
\setlength{\tabcolsep}{4pt}
\renewcommand{\arraystretch}{1.14}
\begin{tabularx}{\linewidth}{@{}lXX@{}}
\toprule
\textbf{Policy} & \textbf{Candidate menu} & \textbf{Decision rule} \\
\midrule
Fixed $h_0$ & Common initial candidate & Evaluate once; retain as the unrevised reference. \\
Flat retry & First 12 unseen one-edit neighbors, sorted by candidate ID & LLM selects an ID from the local menu. \\
Random routing & One canonical edit at a uniformly sampled available address & LLM returns the pre-instantiated ID/address pair. \\
AIDE-style draft & 12 uniformly sampled unseen IDs from the full language & LLM selects an independent root candidate. \\
AIDE-style improve & First 12 unseen neighbors of the best-feedback node, sorted by ID & LLM selects a one-edit improvement. \\
\rowcolor{csaccent}
CellScientist & One canonical edit at each of at most two ranked addresses & LLM selects an allow-listed ID/address pair. \\
\bottomrule
\end{tabularx}
\end{table}

Menus contain fewer than 12 candidates when fewer unseen choices remain. The local parent is the candidate with highest feedback PCC, retaining the earliest observation on ties; the common final retention rule uses mean feedback/selection PCC. CellScientist orders its diagnostic priorities as follows: incomplete input conditioning; direct targets or response PCC exceeding global PCC by more than $0.05$; top-50 response PCC below response PCC; and input/output capacity below 64, or capacity at 64 when the incumbent outperforms an otherwise identical lower-rank history entry. The decoder-error diagnostic can prioritize that address independently of the capacity-based rule. Regularization, target, and conditioning form the fallback priority. Duplicate addresses, exhausted neighborhoods, and addresses already attempted without improvement since the incumbent are removed; the first two remaining addresses form the shortlist. Simultaneous triggers keep this order. The LLM is prompted to prefer the earlier pair unless feedback favors the second; malformed, out-of-menu, or mismatched responses use the first pair. These rules and value preferences are fixed during the reported runs.

\paragraph{Diagnostic schema and graph record.}
The evaluator returns global PCC, sample-wise PCC, MSE, pooled $R^2$ with feature-wise centering, centered PCC, response PCC, response sample-wise PCC, response MSE, and top-20/top-50 response PCC. Responses subtract each row's control profile. Global and response PCC flatten their matrices; top-$K$ PCC averages row-wise correlations over the $K$ largest absolute observed responses. Ranking reads global, response, and top-50 response PCC, plus configuration and history. Fixed dependency edges are conditioning $\to$ representation/target, target $\to$ decoder, representation $\to$ decoder/regularization, and decoder $\to$ regularization. These edges are recorded for inspection; legal transitions are determined by the configuration domains and one-field neighborhoods. For a selected pair $(j,h')$, $\mathcal T_{\mathrm{edit}}$ replaces field $j$ of the incumbent by its value in $h'$. The deterministic preference order is shared with Random routing, so their difference is address selection rather than a different within-address candidate generator.

\paragraph{Executable checks.}\label{app:executable_checks}
Before fitting, the live controlled runner validates domain membership, response allow lists, address--ID pairing, and duplicate IDs. After prediction it checks shape and finite outputs/metrics. Atomic-transition status is recorded with the trial. Proposal/execution exceptions invoke an unseen one-field recovery neighbor, deterministically indexed by seed and step; the common candidate budget remains in force. The separate LCA record audit additionally tests protected-state mutation, exact interface fields, non-atomic edits, route consistency, and repair-budget exhaustion. Its repair actions restore protected fields, remove extra interface fields, or correct the routed address; output-shape and numerical failures are rejected. Thus the registered audit and live-search checks have explicit, distinct scopes.

\paragraph{Language and comparison scope.}
All 768 configurations instantiate the same low-rank fitting pipeline; input blocks and ranks change fitting cost and representational capacity. The controlled contrast evaluates the combined routing policy within this developed design space. Deterministic ranking and LLM pair selection are evaluated jointly; the no-history audit separately measures proposal repetition.

\paragraph{AIDE-style correspondence.}
The adaptation retains the reference algorithm's initial drafting and subsequent greedy improvement of the best valid node~\citep{jiang2025aide}. After the common $h_0$, it creates five new root drafts when budget permits, then improves the best-feedback node; all evaluations count toward $B$. The executable candidate language supplies the configurations for both stages. The reference algorithm's debugging branch is conditional on buggy code, whereas these configurations are executable by construction. The recorded reference is the official implementation at commit \href{https://github.com/WecoAI/aideml/tree/5d66a21771e98623dc9fc8716bdbe388d63464c0}{\texttt{5d66a21}}. This implements its draft-then-improve policy under the common predictor and evaluation budget.

\paragraph{Fixed execution protocol.}
After development, the finalized language and controller code are fixed for the reported five-seed comparisons. Before formal execution, the runner verifies a lock containing the configuration, dataset, and source-file hashes, and rejects a mismatch. Each run records the verified protocol and source-bundle hashes. The reported Gemini and Qwen runs share the same source bundle, including the candidate domains, neighborhood construction, and policy implementations; the backbone-specific configurations have separate locks. This records execution under the finalized protocol.

\FloatBarrier
\subsubsection{LLM Backbones}
\label{app:llm_backbones}

The matched-budget audit uses Gemini 3 Pro for the manuscript-default runs and repeats the controller comparison with the open-weight Qwen2.5-0.5B-Instruct model. Within each backbone, every controller receives the same decoding configuration and candidate budget. Table~\ref{tab:llm_backbone_sources} records the public model identities used in this audit.

\begin{table*}[t]
\centering
\caption{
\textbf{LLM backbones used in the controlled framework audit.}
Gemini 3 Pro is the manuscript-default backbone; Qwen2.5-0.5B-Instruct provides an open-weight reproduction.
}
\label{tab:llm_backbone_sources}
\renewcommand{\arraystretch}{1.02}
\setlength{\tabcolsep}{4pt}
\footnotesize
\begin{tabularx}{\textwidth}{@{}>{\raggedright\arraybackslash}p{0.19\textwidth}>{\raggedright\arraybackslash}p{0.28\textwidth}>{\raggedright\arraybackslash}p{0.18\textwidth}X@{}}
\toprule
\textbf{Main-text Label} & \textbf{Full Backbone Identity} & \textbf{Provider / Family} & \textbf{Source Link} \\
\midrule
Gemini 3 Pro 
& Gemini 3 Pro 
& Google Gemini 
& \url{https://blog.google/products/gemini/gemini-3} \\
Qwen 0.5B
& Qwen2.5-0.5B-Instruct
& Qwen 
& \url{https://huggingface.co/Qwen/Qwen2.5-0.5B-Instruct} \\
\bottomrule
\end{tabularx}
\end{table*}

\paragraph{Controller-call accounting.}
Gemini completed 720 controller decisions with valid allow-listed responses and no fallback. Qwen completed 720 decisions; invalid field pairings triggered the registered legal fallback in 10/180 CellScientist decisions and 70/180 AIDE-style decisions. Flat and Random routing used no fallback. All decisions count toward the reported candidate budgets.

\FloatBarrier
\section{Controller-Backbone Reproduction}
\label{app:backbone_reproduction}

\begin{table}[H]
\centering
\caption{
\textbf{Task-level reproduction of structured routing across controller backbones.}
The upper panel reports mean$\pm$sample SD in held-out global PCC over five paired controller/fitting seeds. The lower panel reports CellScientist-minus-Random effects with descriptive four-setting 95\% $t$ intervals, computed from unrounded results. CS denotes CellScientist. Both backbones use the same task settings, candidate language, fitting, and budgets. Bold marks the better mean within each backbone, setting, and budget.
}
\label{tab:open_backbone_reproduction}
\footnotesize
\renewcommand{\arraystretch}{1.1}
\setlength{\tabcolsep}{3pt}
\renewcommand{\cellres}[2]{\shortstack{$#1$\\{\scriptsize $\pm#2$}}}
\renewcommand{\cellbest}[2]{\shortstack{$\mathbf{#1}$\\{\scriptsize $\pm#2$}}}
\renewcommand{\effectci}[3]{\shortstack{$#1$\\{\scriptsize $[#2,#3]$}}}
\begin{tabular}{lll|cc|cc|cc}
\toprule
\textbf{Backbone}
& \textbf{Dataset} & \textbf{Split}
& \multicolumn{2}{c|}{$B=3$}
& \multicolumn{2}{c|}{$B=5$}
& \multicolumn{2}{c}{$B=10$} \\
\cmidrule(lr){4-5}\cmidrule(lr){6-7}\cmidrule(lr){8-9}
& & & \textbf{Random} & \textbf{CS}
& \textbf{Random} & \textbf{CS}
& \textbf{Random} & \textbf{CS} \\
\midrule
\multirow{4}{*}{Gemini 3 Pro}
& BBBC036 & Plate
& \cellres{0.2250}{0.0493} & \cellbest{0.2650}{0.0005}
& \cellres{0.2277}{0.0502} & \cellbest{0.2674}{0.0003}
& \cellres{0.2626}{0.0077} & \cellbest{0.2693}{0.0012} \\
& BBBC036 & SMILES
& \cellres{0.2052}{0.0413} & \cellbest{0.2336}{0.0016}
& \cellres{0.2049}{0.0417} & \cellbest{0.2346}{0.0009}
& \cellbest{0.2347}{0.0016} & \cellres{0.2346}{0.0009} \\
& BBBC047 & Plate
& \cellres{0.3032}{0.0500} & \cellbest{0.3396}{0.0004}
& \cellres{0.3066}{0.0514} & \cellbest{0.3457}{0.0004}
& \cellres{0.3458}{0.0027} & \cellbest{0.3481}{0.0009} \\
& BBBC047 & SMILES
& \cellres{0.3131}{0.0536} & \cellbest{0.3529}{0.0010}
& \cellres{0.3207}{0.0582} & \cellbest{0.3614}{0.0003}
& \cellres{0.3668}{0.0073} & \cellbest{0.3751}{0.0017} \\
\midrule
\multirow{4}{*}{Qwen2.5-0.5B}
& BBBC036 & Plate
& \cellres{0.2250}{0.0493} & \cellbest{0.2566}{0.0024}
& \cellres{0.2277}{0.0502} & \cellbest{0.2566}{0.0024}
& \cellres{0.2626}{0.0077} & \cellbest{0.2657}{0.0035} \\
& BBBC036 & SMILES
& \cellres{0.2052}{0.0413} & \cellbest{0.2357}{0.0004}
& \cellres{0.2049}{0.0417} & \cellbest{0.2357}{0.0004}
& \cellbest{0.2347}{0.0016} & \cellres{0.2346}{0.0009} \\
& BBBC047 & Plate
& \cellres{0.3032}{0.0500} & \cellbest{0.3429}{0.0003}
& \cellres{0.3066}{0.0514} & \cellbest{0.3425}{0.0009}
& \cellbest{0.3458}{0.0027} & \cellres{0.3438}{0.0018} \\
& BBBC047 & SMILES
& \cellres{0.3131}{0.0536} & \cellbest{0.3570}{0.0005}
& \cellres{0.3207}{0.0582} & \cellbest{0.3615}{0.0005}
& \cellres{0.3668}{0.0073} & \cellbest{0.3707}{0.0002} \\
\midrule
\multicolumn{9}{l}{\textit{CellScientist $-$ Random: mean effect with paired-setting 95\% CI}} \\
\textbf{Backbone}
& \multicolumn{2}{c|}{\textbf{Contrast}}
& \multicolumn{2}{c|}{$B=3$}
& \multicolumn{2}{c|}{$B=5$}
& \multicolumn{2}{c}{$B=10$} \\
Gemini 3 Pro
& \multicolumn{2}{c|}{CellScientist $-$ Random}
& \multicolumn{2}{c|}{\effectci{+0.0361}{+0.0275}{+0.0448}}
& \multicolumn{2}{c|}{\effectci{+0.0373}{+0.0292}{+0.0454}}
& \multicolumn{2}{c}{\effectci{+0.0043}{-0.0018}{+0.0104}} \\
Qwen2.5-0.5B
& \multicolumn{2}{c|}{CellScientist $-$ Random}
& \multicolumn{2}{c|}{\effectci{+0.0364}{+0.0261}{+0.0467}}
& \multicolumn{2}{c|}{\effectci{+0.0341}{+0.0256}{+0.0426}}
& \multicolumn{2}{c}{\effectci{+0.0012}{-0.0032}{+0.0056}} \\
\bottomrule
\end{tabular}
\end{table}

\input{existing_run_statistics}
\FloatBarrier
\section{Detailed LCA and PDR Audit Results}
\label{app:lca_pdr_audits}

Table~\ref{tab:mechanism_audit_summary} summarizes the HRT, LCA, and PDR results in the main text. Here we report LCA outcomes and PDR results by component address, including registered case counts, decisions, localization, repair success, unrelated edits, and repair rounds.

\subsection{Registered LCA Outcome Audit}

The LCA audit constructs 40 single-violation request records and 20 legal controls. The validator repairs 15 records, blocks 15 at the pre-run gate, and rejects 10 injected runtime-output records. Legal controls remain unchanged, yielding 60/60 correct decisions on these registered cases. The 30 pre-run interventions count requests corrected or stopped before dispatch in this audit, rather than timed model-training jobs. Its registered families test the implemented checker predicates.

\begin{table}[htpb]
\centering
\caption{
\textbf{Registered LCA audit grouped by outcome.}
Each positive request record contains one injected violation; legal controls contain none. ``Eligible'' denotes acceptance after record repair, and ``Avoided'' counts requests corrected or stopped at the pre-run gate. Runtime records are injected for post-run checks. Across these registered cases, violation recall, exact-code accuracy, legal-case specificity, and repair-or-safe-block containment are all 100\%.
}
\label{tab:lca_registered_audit}
\scriptsize
\renewcommand{\arraystretch}{1.02}
\setlength{\tabcolsep}{2pt}
\begin{tabular}{@{}
>{\raggedright\arraybackslash}p{0.15\linewidth}
>{\raggedright\arraybackslash}p{0.33\linewidth}
>{\centering\arraybackslash}p{0.045\linewidth}
>{\centering\arraybackslash}p{0.045\linewidth}
>{\centering\arraybackslash}p{0.090\linewidth}
>{\centering\arraybackslash}p{0.085\linewidth}
>{\centering\arraybackslash}p{0.065\linewidth}
>{\centering\arraybackslash}p{0.060\linewidth}@{}}
\toprule
\textbf{LCA outcome}
& \textbf{Registered cases}
& \textbf{Stage}
& \textbf{Cases}
& \shortstack{\textbf{Detected /}\\\textbf{accepted}}
& \shortstack{\textbf{Observed}\\\textbf{action}}
& \textbf{Eligible}
& \textbf{Avoided} \\
\midrule
Repair
& Protected-state mutation; interface mismatch; route mismatch
& Pre & 15 & \textbf{15/15 detected} & \textbf{15/15 repaired} & \textbf{15/15} & 15 \\
Pre-run block
& Out-of-domain candidate; non-atomic edit; repair budget exhausted
& Pre & 15 & \textbf{15/15 detected} & \textbf{15/15 blocked} & 0/15 & 15 \\
Post-run block
& Output-shape mismatch; non-finite output
& Post & 10 & \textbf{10/10 detected} & \textbf{10/10 blocked} & 0/10 & 0 \\
Legal controls
& Contract-preserving candidates across registered interfaces and component domains
& Pre & 20 & \textbf{20/20 accepted} & Unchanged & \textbf{20/20} & 0 \\
\midrule
\textbf{Overall}
& \multicolumn{2}{l}{All registered violation and legal-control cases}
& \textbf{60}
& \textbf{60/60 correct}
& \textbf{40/40 contained}
& \textbf{35/60 eligible}
& \textbf{30} \\
\bottomrule
\end{tabular}
\end{table}

\subsection{Held-Out Registered-Fault PDR Audit}

The PDR audit constructs four variants of each of five typed-state fault families, with one variant per family assigned to calibration and three to reporting. The controller receives a candidate configuration and constructed metric values; the diagnostic label is retained for scoring. Repair is the restoration of the registered configuration condition, checked without refitting: required conditioning, response target, input/output capacity, or regularization. Flat retry, Random routing, and structured routing receive the same cases. Structured routing restores every held-out state with 0 unrelated edits and 1.00 mean round. These are localization and state-restoration measurements within known fault families.

\begin{table}[htpb]
\centering
\caption{
\textbf{Held-out registered-fault PDR audit by component address.}
Five injected typed-state cases are used for calibration and 15 for reporting, with three reporting variants per address. Structured PDR uses Gemini 3 Pro to select an allow-listed pair from candidate-state and metric feedback. Repair denotes restoration of the registered field condition. The repair budget is five rounds; displayed means assign five rounds to unrepaired cases. Best values within each address are \textbf{bolded}; ties share emphasis.
}
\label{tab:pdr_registered_audit}
\renewcommand{\arraystretch}{0.92}
\setlength{\tabcolsep}{4pt}
\resizebox{\linewidth}{!}{%
\begin{tabular}{llcccccc}
\toprule
\textbf{Routing policy}
& \textbf{Expected address}
& \textbf{Cases}
& \textbf{Top-1 accuracy $\uparrow$}
& \textbf{Top-2 recall $\uparrow$}
& \textbf{Repair success $\uparrow$}
& \textbf{Mean unrelated edits $\downarrow$}
& \textbf{Mean repair rounds $\downarrow$} \\
\midrule
\multirow{6}{*}{Flat retry}
& Conditioning & 3 & 66.7\% & N/A & 0.0\% & 4.33 & 5.00 \\
& Response parameterization & 3 & 0.0\% & N/A & \textbf{100\%} & 1.00 & 2.00 \\
& Representation & 3 & 0.0\% & N/A & 0.0\% & 4.00 & 5.00 \\
& Decoder & 3 & 0.0\% & N/A & 0.0\% & 4.33 & 5.00 \\
& Regularization & 3 & 0.0\% & N/A & 66.7\% & 2.33 & 3.67 \\
& All held-out & 15 & 13.3\% & N/A & 33.3\% & 3.20 & 4.13 \\
\midrule
\multirow{6}{*}{Random routing}
& Conditioning & 3 & 0.0\% & 0.0\% & 0.0\% & 5.00 & 5.00 \\
& Response parameterization & 3 & 33.3\% & 33.3\% & 66.7\% & 2.67 & 3.33 \\
& Representation & 3 & 0.0\% & 0.0\% & 66.7\% & 3.33 & 4.00 \\
& Decoder & 3 & 0.0\% & 33.3\% & \textbf{100\%} & 2.33 & 3.33 \\
& Regularization & 3 & 0.0\% & 0.0\% & 66.7\% & 3.00 & 3.67 \\
& All held-out & 15 & 6.7\% & 13.3\% & 60.0\% & 3.27 & 3.87 \\
\midrule
\multirow{6}{*}{Structured PDR}
& Conditioning & 3 & \textbf{100\%} & \textbf{100\%} & \textbf{100\%} & \textbf{0.00} & \textbf{1.00} \\
& Response parameterization & 3 & \textbf{100\%} & \textbf{100\%} & \textbf{100\%} & \textbf{0.00} & \textbf{1.00} \\
& Representation & 3 & \textbf{100\%} & \textbf{100\%} & \textbf{100\%} & \textbf{0.00} & \textbf{1.00} \\
& Decoder & 3 & \textbf{100\%} & \textbf{100\%} & \textbf{100\%} & \textbf{0.00} & \textbf{1.00} \\
& Regularization & 3 & \textbf{100\%} & \textbf{100\%} & \textbf{100\%} & \textbf{0.00} & \textbf{1.00} \\
& All held-out & 15 & \textbf{100\%} & \textbf{100\%} & \textbf{100\%} & \textbf{0.00} & \textbf{1.00} \\
\bottomrule
\end{tabular}
}
\end{table}

\section{Additional Experimental Analysis}
\label{app:detailed_experiments}

This appendix characterizes the high-freedom workflow, evaluates recovery of responsive morphology features, and reports additional single-cell results. These analyses complement the main results under their respective evaluation protocols.

\subsection{Validation-Selected Discovery Outcomes}
\label{app:discovery_outcomes}

\begin{figure}[!t]
\centering
\begin{minipage}[t]{0.98\textwidth}
    \centering
    \begin{minipage}[t]{0.485\linewidth}
        \centering
        \includegraphics[width=\linewidth]{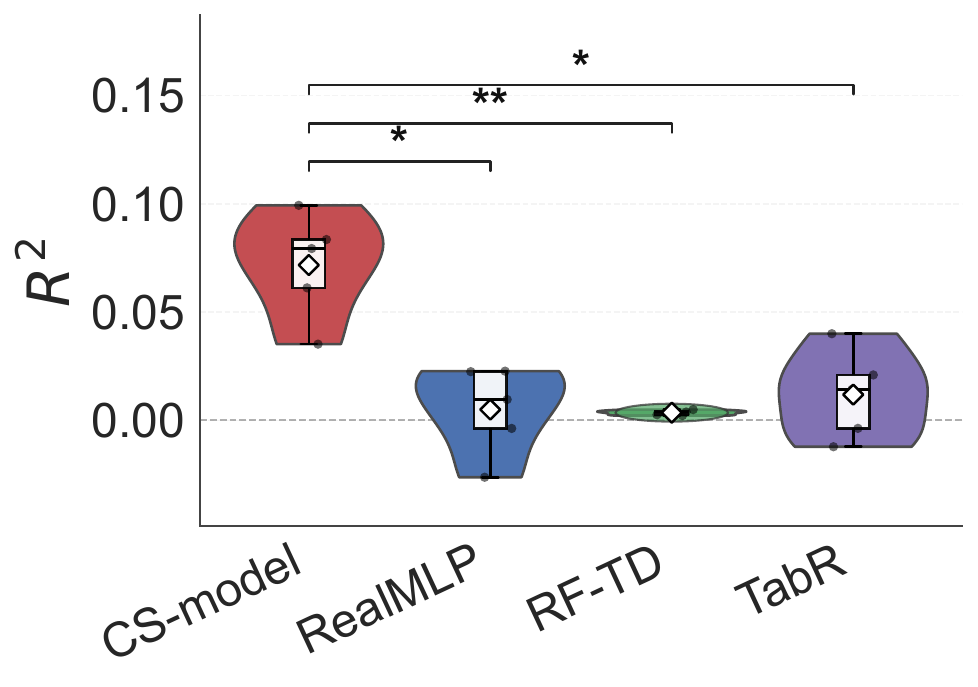}
        \par\footnotesize (a) BBBC036.
    \end{minipage}\hfill
    \begin{minipage}[t]{0.485\linewidth}
        \centering
        \includegraphics[width=\linewidth]{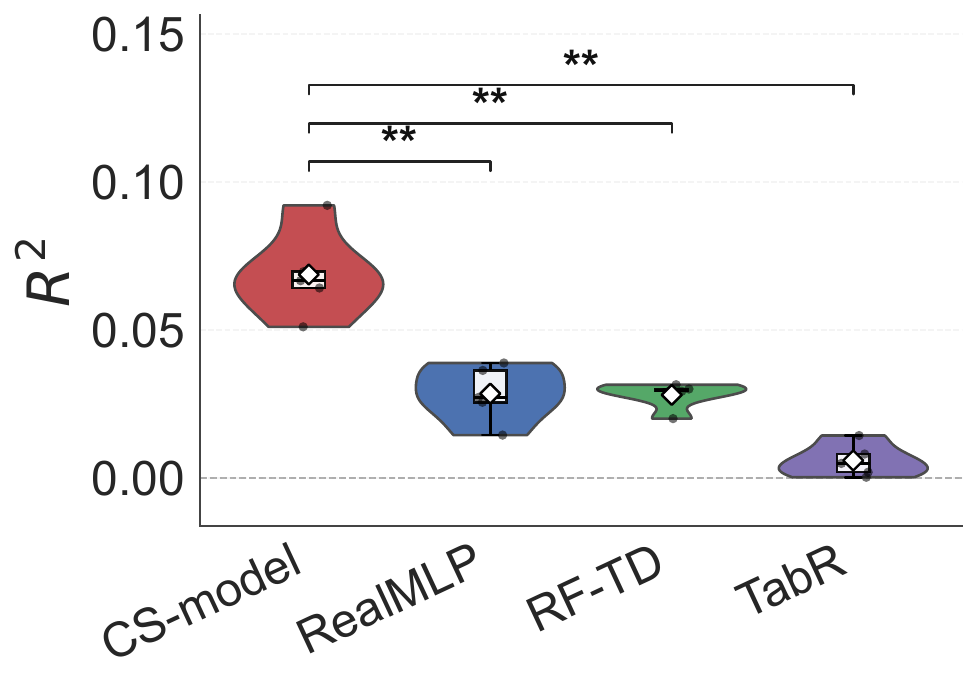}
        \par\footnotesize (b) BBBC047.
    \end{minipage}
    \captionof{figure}{
    \textbf{Fold-level comparison on BBBC036 and BBBC047.}
    Violins show five-fold $R^2$; boxes, points, and diamonds denote quartiles, folds, and means. Stars: uncorrected paired $t$-tests, $^{*}p<0.05$, $^{**}p<0.01$, $^{***}p<0.001$.
    }
    \label{fig:r2_fold_comparison}
\end{minipage}
\end{figure}

The high-freedom workflow selects task-specific predictors for morphology and transcriptomic response tasks. Table~\ref{tab:sota_benchmark} reports five-fold validation mean and sample SD under exact-SMILES and plate grouping. CS-model has the best mean MSE, PCC, and $R^2$ among the listed methods in these comparisons. Figure~\ref{fig:r2_fold_comparison} shows the corresponding fold-level $R^2$ values and paired-test markers; response-feature metrics follow in Appendix~\ref{app:morphology_results}.

The morphology baselines are TabR~\citep{gorishniy2024tabr}, RealMLP and RF-TD~\citep{holzmuller2024better}, FLAML~\citep{wang2021flaml}, Random Search~\citep{gijsbers2024amlb}, and CellForge~\citep{tang2025cellforge}. In addition to BBBC036 and BBBC047, CPG0016 extends the morphology search to a larger compound collection~\citep{chandrasekaran2023jump,arevalo2024evaluating}.

\begin{table}[!t]
\centering
\caption{
\textbf{Validation-selected morphology-response models.}
CS-model denotes the task-specific predictor selected by CellScientist. Entries are five-fold validation mean$\pm$sample SD under SMILES- and plate-based splits. Blue shading identifies CellScientist models; \textbf{bold} marks the best mean within each comparison.
}
\label{tab:sota_benchmark}
\scriptsize
\setlength{\tabcolsep}{2.4pt}
\renewcommand{\arraystretch}{0.5}
\resizebox{\linewidth}{!}{%
\begin{tabular}{c c|ccc|ccc}
\toprule
\multirow{2}{*}{\textbf{Method}} & \multirow{2}{*}{\textbf{Venue}}
& \multicolumn{3}{c|}{\textbf{SMILES-based Split}} 
& \multicolumn{3}{c}{\textbf{Plate-based Split}} \\
\cmidrule(lr){3-5} \cmidrule(lr){6-8}
& & \textbf{MSE}$\downarrow$ & \textbf{PCC}$\uparrow$ & \textbf{R$^2$}$\uparrow$
& \textbf{MSE}$\downarrow$ & \textbf{PCC}$\uparrow$ & \textbf{R$^2$}$\uparrow$ \\
\midrule
\multicolumn{8}{c}{\textbf{BBBC036}} \\
\midrule
\multicolumn{8}{l}{\textbf{$\#$ Existing reference methods}} \\
TabR & ICLR'24
& \cellres{3.5518}{0.24} & \cellres{0.2747}{0.03} & \cellres{0.0119}{0.02}
& \cellres{2.8759}{0.13} & \cellres{0.5065}{0.03} & \cellres{0.1855}{0.02} \\
RealMLP & NeurIPS'24
& \cellres{3.5305}{0.16} & \cellres{0.3301}{0.01} & \cellres{0.0049}{0.02}
& \cellres{2.5740}{0.16} & \cellres{0.5856}{0.02} & \cellres{0.2530}{0.03} \\
RF-TD & NeurIPS'24
& \cellres{3.5951}{0.19} & \cellres{0.2447}{0.01} & \cellres{0.0035}{0.01}
& \cellres{3.5978}{0.14} & \cellres{0.2466}{0.02} & \cellres{0.0152}{0.01} \\
\midrule
\multicolumn{8}{l}{\textbf{$\square$ Search-based baselines}} \\
FLAML & MLSys'21
& \cellres{3.4217}{0.18} & \cellres{0.3486}{0.02} & \cellres{0.0379}{0.02}
& \cellres{2.4918}{0.14} & \cellres{0.5987}{0.02} & \cellres{0.2784}{0.03} \\
Random Search & JMLR'24
& \cellres{3.4763}{0.21} & \cellres{0.3392}{0.02} & \cellres{0.0246}{0.02}
& \cellres{2.5335}{0.15} & \cellres{0.5904}{0.02} & \cellres{0.2649}{0.03} \\
\midrule
\multicolumn{8}{l}{\textbf{$\triangle$ Agent-designed methods}} \\
CellForge & arXiv'25
& \cellres{3.4568}{0.19} & \cellres{0.3415}{0.03} & \cellres{0.0464}{0.02}
& \cellres{2.5187}{0.14} & \cellres{0.6013}{0.02} & \cellres{0.2878}{0.03} \\
\rowcolor{csaccent}
\textbf{CS-model}& Ours
& \cellbest{3.3115}{0.20} & \cellbest{0.3688}{0.03} & \cellbest{0.0717}{0.02}
& \cellbest{2.3662}{0.15} & \cellbest{0.6212}{0.02} & \cellbest{0.3165}{0.03} \\
\midrule
\multicolumn{8}{c}{\textbf{BBBC047}} \\
\midrule
\multicolumn{8}{l}{\textbf{$\#$ Existing reference methods}} \\
TabR & ICLR'24
& \cellres{2.9382}{0.04} & \cellres{0.3552}{0.01} & \cellres{0.0059}{0.01}
& \cellres{2.7930}{0.14} & \cellres{0.4094}{0.01} & \cellres{0.0568}{0.03} \\
RealMLP & NeurIPS'24
& \cellres{2.8441}{0.03} & \cellres{0.3985}{0.01} & \cellres{0.0285}{0.01}
& \cellres{2.6475}{0.10} & \cellres{0.4597}{0.01} & \cellres{0.1332}{0.02} \\
RF-TD & NeurIPS'24
& \cellres{2.8850}{0.05} & \cellres{0.3746}{0.01} & \cellres{0.0281}{0.01}
& \cellres{2.8174}{0.16} & \cellres{0.4005}{0.01} & \cellres{0.0615}{0.01} \\
\midrule
\multicolumn{8}{l}{\textbf{$\square$ Search-based baselines}} \\
FLAML & MLSys'21
& \cellres{2.8126}{0.06} & \cellres{0.4127}{0.01} & \cellres{0.0435}{0.01}
& \cellres{2.6158}{0.11} & \cellres{0.4698}{0.01} & \cellres{0.1407}{0.02} \\
Random Search & JMLR'24
& \cellres{2.8314}{0.07} & \cellres{0.4059}{0.01} & \cellres{0.0358}{0.01}
& \cellres{2.6342}{0.12} & \cellres{0.4635}{0.01} & \cellres{0.1362}{0.02} \\
\midrule
\multicolumn{8}{l}{\textbf{$\triangle$ Agent-designed methods}} \\
CellForge & arXiv'25
& \cellres{2.7869}{0.09} & \cellres{0.4154}{0.01} & \cellres{0.0476}{0.01}
& \cellres{2.5908}{0.10} & \cellres{0.4772}{0.02} & \cellres{0.1471}{0.02} \\
\rowcolor{csaccent}
\textbf{CS-model}& Ours
& \cellbest{2.7612}{0.11} & \cellbest{0.4368}{0.01} & \cellbest{0.0687}{0.01}
& \cellbest{2.5704}{0.11} & \cellbest{0.4844}{0.02} & \cellbest{0.1538}{0.03} \\
\midrule
\multicolumn{8}{c}{\textbf{CPG0016}} \\
\midrule
\multicolumn{8}{l}{\textbf{$\#$ Existing reference methods}} \\
TabR & ICLR'24
& \cellres{2.2622}{0.12} & \cellres{0.0770}{0.01} & \cellres{-0.5516}{0.04}
& \cellres{2.0667}{0.04} & \cellres{0.1953}{0.01} & \cellres{-0.4128}{0.02} \\
RealMLP & NeurIPS'24
& \cellres{1.4174}{0.08} & \cellres{0.2616}{0.02} & \cellres{0.0297}{0.01}
& \cellres{1.2105}{0.01} & \cellres{0.4418}{0.01} & \cellres{0.1717}{0.01} \\
RF-TD & NeurIPS'24
& \cellres{1.4673}{0.09} & \cellres{0.1354}{0.01} & \cellres{-0.0035}{0.01}
& \cellres{1.3400}{0.01} & \cellres{0.3213}{0.01} & \cellres{0.0849}{0.01} \\
\midrule
\multicolumn{8}{l}{\textbf{$\square$ Search-based baselines}} \\
FLAML & MLSys'21
& \cellres{1.2865}{0.05} & \cellres{0.3748}{0.01} & \cellres{0.1183}{0.01}
& \cellres{1.1206}{0.03} & \cellres{0.5037}{0.02} & \cellres{0.2331}{0.02} \\
Random Search & JMLR'24
& \cellres{1.3428}{0.06} & \cellres{0.3189}{0.02} & \cellres{0.0747}{0.02}
& \cellres{1.1674}{0.03} & \cellres{0.4668}{0.02} & \cellres{0.1994}{0.02} \\
\midrule
\multicolumn{8}{l}{\textbf{$\triangle$ Agent-designed methods}} \\
CellForge & arXiv'25
& \cellres{1.2849}{0.03} & \cellres{0.3716}{0.01} & \cellres{0.1272}{0.01}
& \cellres{1.0458}{0.02} & \cellres{0.5187}{0.03} & \cellres{0.2376}{0.03} \\
\rowcolor{csaccent}
\textbf{CS-model}& Ours
& \cellbest{1.1113}{0.01} & \cellbest{0.5093}{0.01} & \cellbest{0.2383}{0.01}
& \cellbest{0.9937}{0.02} & \cellbest{0.5761}{0.04} & \cellbest{0.3117}{0.04} \\
\bottomrule
\end{tabular}
}
\end{table}

\paragraph{Additional response spaces.}
Table~\ref{tab:lincs_full} reports task-specific searches on the Seven Cell Lines and Full Data settings of LINCS2020 L1000~\citep{subramanian2017high,xie2022getting}. The listed predictors include DLEPS~\citep{zhu2021dleps}, DeepCE~\citep{pham2021deepce}, CIGER~\citep{pham2022ciger}, MultiDCP~\citep{wu2022multidcp}, and TranSiGen~\citep{tong2024transigen}. Metrics cover RMSE, PCC, and positive/negative P@100. Appendix~\ref{app:res_trans} gives results for the Norman and Papalexi single-cell benchmarks~\citep{norman2019exploring,papalexi2021characterizing}, including separate Papalexi RNA and protein readouts. Each task uses its own search and training run.

\begin{table}[!ht]

\centering
\footnotesize

\caption{
\textbf{LINCS2020 transcriptomic perturbation prediction.}
\textbf{CS-model} achieves the best mean results among the listed methods in both the Seven Cell Lines and Full Data settings. Each uses an independent task-specific search. Best means are \textbf{bolded}.
}
\label{tab:lincs_full}
\renewcommand{\arraystretch}{1.0}
\setlength{\tabcolsep}{4pt}
\begin{adjustbox}{max width=\linewidth}
\begin{tabular}{@{}llcccc@{}}
\toprule
\textbf{Dataset} & \textbf{Model} 
& \textbf{RMSE} $\downarrow$ 
& \textbf{PCC} $\uparrow$ 
& \shortstack{\textbf{Positive}\\\textbf{P@100} $\uparrow$}
& \shortstack{\textbf{Negative}\\\textbf{P@100} $\uparrow$} \\
\midrule
\multirow{6}{*}{\textbf{Seven Cell}}
& DLEPS      
& \cellres{1.4954}{0.04} 
& \cellres{0.4268}{0.02} 
& \cellres{0.2460}{0.01} 
& \cellres{0.3087}{0.01} \\
& DeepCE     
& \cellres{1.7728}{0.03} 
& \cellres{0.4121}{0.01} 
& \cellres{0.2282}{0.02} 
& \cellres{0.2713}{0.02} \\
& CIGER      
& \cellres{4.4646}{1.06} 
& \cellres{0.4165}{0.01} 
& \cellres{0.2337}{0.02} 
& \cellres{0.2849}{0.02} \\
& MultiDCP   
& \cellres{1.7697}{0.03} 
& \cellres{0.4461}{0.01} 
& \cellres{0.2378}{0.02} 
& \cellres{0.2872}{0.01} \\
& TranSiGen  
& \cellres{0.6903}{0.02} 
& \cellres{0.5332}{0.02} 
& \cellres{0.3785}{0.01} 
& \cellres{0.3886}{0.02} \\
\rowcolor{csaccent}
& \textbf{CS-model}
& \cellbest{0.5088}{0.02} 
& \cellbest{0.6183}{0.02} 
& \cellbest{0.4450}{0.02} 
& \cellbest{0.4481}{0.02} \\
\midrule
& TranSiGen  
& \cellres{0.5186}{0.01} 
& \cellres{0.6076}{0.01} 
& \cellres{0.4353}{0.01} 
& \cellres{0.4403}{0.01} \\
\rowcolor{csaccent}
\multirow{-2}{*}{\textbf{Full Data}} & \textbf{CS-model}
& \cellbest{0.4967}{0.01} 
& \cellbest{0.6368}{0.01} 
& \cellbest{0.4583}{0.01} 
& \cellbest{0.4652}{0.01} \\
\bottomrule
\end{tabular}
\end{adjustbox}

\end{table}
\FloatBarrier

\subsection{End-to-End Open-Workflow Characterization}

\textbf{Open-workflow evidence.}
Tables~\ref{tab:sota_benchmark} and~\ref{tab:lincs_full} report task-specific predictors selected by the high-freedom CellScientist workflow under morphology and transcriptomic protocols. Figures~\ref{fig:r2_fold_comparison} and~\ref{fig:open_workflow_trajectory} show fold-level comparisons and non-monotonic revision trajectories, respectively. These results assess predictor quality; the main-text RQ1 comparison separately compares controller policies within a fixed candidate language.

\label{app:open_workflow_characterization}

The RQ1 study compares controller policies within a fixed executable language. Table~\ref{tab:open_workflow_characterization} complements that audit with an end-to-end comparison in which implementations and selected model families can vary. Cumulative configurations share the task specification, split, validation metric, iteration budget, repair budget, execution timeout, and environment. The comparison records execution success, rule-compliant success, predictive quality, and runtime as structured task state, configured checks, and history-conditioned revision are added.

\begin{table*}[ht]
\centering
\caption{
\textbf{End-to-end open-workflow characterization.}
SR denotes execution success, R-SR denotes rule-compliant success, and Avg/Best PCC summarize validation performance over eligible runs. Time reports wall-clock runtime. This high-freedom characterization complements the fixed-language comparisons in RQ1.
}
\label{tab:open_workflow_characterization}
\renewcommand{\arraystretch}{1}
\setlength{\tabcolsep}{4pt}
\resizebox{\linewidth}{!}{%
\begin{tabular}{cc|ccccc|ccccc}
\toprule
\multirow{2}{*}{\textbf{ID}} & \multirow{2}{*}{\textbf{Method Configuration}} & \multicolumn{5}{c|}{\textbf{BBBC036 (SMILES-based Split)}} & \multicolumn{5}{c}{\textbf{BBBC047 (SMILES-based Split)}} \\
\cmidrule(lr){3-7} \cmidrule(lr){8-12}
 & & \textbf{SR}$\uparrow$ & \textbf{R-SR}$\uparrow$ & \textbf{Avg PCC}$\uparrow$ & \textbf{Best PCC}$\uparrow$ & \textbf{Time (h)}$\downarrow$ & \textbf{SR}$\uparrow$ & \textbf{R-SR}$\uparrow$ & \textbf{Avg PCC}$\uparrow$ & \textbf{Best PCC}$\uparrow$ & \textbf{Time (h)}$\downarrow$ \\
\midrule
M0 & Prompt + LLM + Code Executor & 0.526 & 0.526 & 0.1357 & 0.1682 & 10.25 & 0.368 & 0.368 & 0.1422 & 0.1752 & 15.58 \\
M1 & M0 + Re-prompting (Naive) & 0.263 & 0.263 & 0.2156 & 0.2485 & 17.96 & 0.210 & 0.210 & 0.2267 & 0.2581 & 27.29 \\
M2 & M1 + Structured Task State (HRT) & 0.947 & 0.684 & 0.2771 & 0.3050 & 10.34 & 0.941 & 0.588 & 0.2472 & 0.2775 & 15.96 \\
M3 & M2 + Protected Editing / Checks (LCA) & \textbf{1.000} & 0.737 & 0.3183 & 0.3214 & 8.46 & \textbf{1.000} & 0.684 & 0.3315 & 0.3962 & 13.38 \\
\textbf{M4} & \textbf{M3 + History-Conditioned Revision (PDR)} & \textbf{1.000} & \textbf{0.789} & \textbf{0.3413} & \textbf{0.3688} & \textbf{7.93} & \textbf{1.000} & \textbf{0.714} & \textbf{0.3539} & \textbf{0.4368} & \textbf{12.58} \\
\bottomrule
\end{tabular}%
}
\end{table*}

On both datasets, M4 achieved the highest R-SR, Avg PCC, and Best PCC, and the lowest runtime among the tested open-workflow configurations. Figure~\ref{fig:open_workflow_trajectory} records exploratory regressions and validation-retained improvements, linking these aggregate results to the case-study trajectories.

\subsection{Response-sensitive Morphology Metrics}
\label{app:morphology_results}

Table~\ref{tab:appendix_deg_benchmark} describes the observed method differences on the most perturbation-responsive morphology features. We report response-feature RMSE and PCC over the top-$K$ responsive features on BBBC036, BBBC047, and CPG0016 under both SMILES-based and plate-based splits.

\begin{table*}[t]
\centering
\caption{
\textbf{Response-sensitive morphology metrics.}
Response-feature metrics evaluate prediction quality on the top-$K$ most responsive morphology features. CS-model has the best mean values among the listed methods under both SMILES-based and plate-based splits in this table. Best mean values are \textbf{bolded}.
}
\label{tab:appendix_deg_benchmark}
\setlength{\tabcolsep}{2.5pt}
\renewcommand{\arraystretch}{0.90}
\resizebox{\linewidth}{!}{%
\begin{tabular}{c c|cccc|cccc}
\toprule
\multirow{2}{*}{\textbf{Method}} & \multirow{2}{*}{\textbf{Venue}}
& \multicolumn{4}{c|}{\textbf{SMILES-based Split}} 
& \multicolumn{4}{c}{\textbf{Plate-based Split}} \\
\cmidrule(lr){3-6} \cmidrule(lr){7-10}
& & 
\textbf{$\text{RMSE}_{\text{RF-20}}$}$\downarrow$ 
& \textbf{$\text{RMSE}_{\text{RF-50}}$}$\downarrow$ 
& \textbf{$\text{PCC}_{\text{RF-20}}$}$\uparrow$ 
& \textbf{$\text{PCC}_{\text{RF-50}}$}$\uparrow$
& \textbf{$\text{RMSE}_{\text{RF-20}}$}$\downarrow$ 
& \textbf{$\text{RMSE}_{\text{RF-50}}$}$\downarrow$ 
& \textbf{$\text{PCC}_{\text{RF-20}}$}$\uparrow$ 
& \textbf{$\text{PCC}_{\text{RF-50}}$}$\uparrow$ \\
\midrule

\multicolumn{10}{c}{\textbf{BBBC036}} \\
\midrule
\multicolumn{10}{l}{\textbf{$\#$ Existing reference methods}} \\
TabR & ICLR'24
& \cellres{4.1306}{0.03} & \cellres{3.3883}{0.04} & \cellres{0.2742}{0.01} & \cellres{0.2584}{0.01}
& \cellres{3.8974}{0.12} & \cellres{3.1610}{0.07} & \cellres{0.3784}{0.05} & \cellres{0.3783}{0.05} \\
RealMLP & NeurIPS'24
& \cellres{3.9910}{0.03} & \cellres{3.2980}{0.03} & \cellres{0.3170}{0.01} & \cellres{0.2929}{0.01}
& \cellres{3.5303}{0.13} & \cellres{2.8547}{0.08} & \cellres{0.4636}{0.03} & \cellres{0.4681}{0.02} \\
RF-TD & NeurIPS'24
& \cellres{4.1499}{0.02} & \cellres{3.4356}{0.03} & \cellres{0.2445}{0.01} & \cellres{0.2227}{0.01}
& \cellres{3.9590}{0.13} & \cellres{3.2927}{0.07} & \cellres{0.5268}{0.05} & \cellres{0.4649}{0.05} \\
\midrule
\multicolumn{10}{l}{\textbf{$\square$ Search-based baselines}} \\
FLAML & MLSys'21
& \cellres{3.9892}{0.04} & \cellres{3.2897}{0.04} & \cellres{0.3297}{0.01} & \cellres{0.3079}{0.01}
& \cellres{3.3658}{0.11} & \cellres{2.7702}{0.08} & \cellres{0.6486}{0.03} & \cellres{0.6705}{0.03} \\
Random Search & JMLR'24
& \cellres{3.9948}{0.05} & \cellres{3.2964}{0.05} & \cellres{0.3218}{0.01} & \cellres{0.2978}{0.01}
& \cellres{3.4316}{0.12} & \cellres{2.8357}{0.08} & \cellres{0.6235}{0.04} & \cellres{0.6468}{0.04} \\
\midrule
\multicolumn{10}{l}{\textbf{$\triangle$ Agent-designed methods}} \\
CellForge & arXiv'25
& \cellres{3.9903}{0.05} & \cellres{3.2928}{0.05} & \cellres{0.3262}{0.01} & \cellres{0.3034}{0.01}
& \cellres{3.3869}{0.10} & \cellres{2.7883}{0.08} & \cellres{0.6574}{0.03} & \cellres{0.6813}{0.03} \\
\textbf{CS-model} & Ours
& \cellbest{3.9865}{0.09} & \cellbest{3.2849}{0.08} & \cellbest{0.3339}{0.01} & \cellbest{0.3139}{0.01}
& \cellbest{3.2817}{0.11} & \cellbest{2.7046}{0.09} & \cellbest{0.6856}{0.02} & \cellbest{0.7082}{0.02} \\

\midrule
\multicolumn{10}{c}{\textbf{BBBC047}} \\
\midrule
\multicolumn{10}{l}{\textbf{$\#$ Existing reference methods}} \\
TabR & ICLR'24
& \cellres{6.2132}{0.02} & \cellres{4.6263}{0.01} & \cellres{0.3355}{0.01} & \cellres{0.3898}{0.01}
& \cellres{6.2120}{0.05} & \cellres{4.6035}{0.04} & \cellres{0.3398}{0.01} & \cellres{0.4009}{0.01} \\
RealMLP & NeurIPS'24
& \cellres{5.8479}{0.02} & \cellres{4.4168}{0.01} & \cellres{0.3678}{0.01} & \cellres{0.4851}{0.01}
& \cellres{6.1791}{0.06} & \cellres{4.5301}{0.04} & \cellres{0.3836}{0.02} & \cellres{0.4270}{0.01} \\
RF-TD & NeurIPS'24
& \cellres{6.0601}{0.03} & \cellres{4.5652}{0.02} & \cellres{0.3742}{0.01} & \cellres{0.4153}{0.01}
& \cellres{6.0934}{0.05} & \cellres{4.5715}{0.04} & \cellres{0.3591}{0.02} & \cellres{0.4063}{0.01} \\
\midrule
\multicolumn{10}{l}{\textbf{$\square$ Search-based baselines}} \\
FLAML & MLSys'21
& \cellres{5.2847}{0.04} & \cellres{4.0209}{0.03} & \cellres{0.3685}{0.01} & \cellres{0.4979}{0.01}
& \cellres{5.5128}{0.05} & \cellres{4.0879}{0.04} & \cellres{0.3892}{0.02} & \cellres{0.5096}{0.01} \\
Random Search & JMLR'24
& \cellres{5.4362}{0.05} & \cellres{4.1328}{0.04} & \cellres{0.3601}{0.01} & \cellres{0.4887}{0.01}
& \cellres{5.6315}{0.06} & \cellres{4.1746}{0.04} & \cellres{0.3838}{0.02} & \cellres{0.4983}{0.01} \\
\midrule
\multicolumn{10}{l}{\textbf{$\triangle$ Agent-designed methods}} \\
CellForge & arXiv'25
& \cellres{5.1164}{0.04} & \cellres{3.8946}{0.03} & \cellres{0.3748}{0.01} & \cellres{0.5096}{0.01}
& \cellres{5.3846}{0.05} & \cellres{3.9907}{0.04} & \cellres{0.3967}{0.02} & \cellres{0.5208}{0.01} \\
\textbf{CS-model} & Ours
& \cellbest{4.9589}{0.03} & \cellbest{3.7581}{0.03} & \cellbest{0.3796}{0.01} & \cellbest{0.5172}{0.01}
& \cellbest{5.2274}{0.04} & \cellbest{3.8788}{0.04} & \cellbest{0.4025}{0.02} & \cellbest{0.5312}{0.01} \\

\midrule
\multicolumn{10}{c}{\textbf{CPG0016}} \\
\midrule
\multicolumn{10}{l}{\textbf{$\#$ Existing reference methods}} \\
TabR & ICLR'24
& \cellres{2.6335}{0.05} & \cellres{2.3924}{0.05} & \cellres{0.0896}{0.01} & \cellres{0.0865}{0.01}
& \cellres{2.5259}{0.01} & \cellres{2.2874}{0.01} & \cellres{0.1330}{0.01} & \cellres{0.1309}{0.01} \\
RealMLP & NeurIPS'24
& \cellres{2.5152}{0.04} & \cellres{2.2456}{0.04} & \cellres{0.3776}{0.02} & \cellres{0.3579}{0.02}
& \cellres{2.3456}{0.01} & \cellres{2.0846}{0.01} & \cellres{0.4658}{0.01} & \cellres{0.4504}{0.01} \\
RF-TD & NeurIPS'24
& \cellres{2.5885}{0.05} & \cellres{2.3130}{0.04} & \cellres{0.0170}{0.01} & \cellres{0.0224}{0.01}
& \cellres{2.4955}{0.01} & \cellres{2.2221}{0.01} & \cellres{0.0758}{0.01} & \cellres{0.0835}{0.01} \\
\midrule
\multicolumn{10}{l}{\textbf{$\square$ Search-based baselines}} \\
FLAML & MLSys'21
& \cellres{2.1126}{0.03} & \cellres{1.9075}{0.03} & \cellres{0.5062}{0.01} & \cellres{0.4974}{0.01}
& \cellres{2.0249}{0.02} & \cellres{1.8186}{0.02} & \cellres{0.5614}{0.02} & \cellres{0.5537}{0.02} \\
Random Search & JMLR'24
& \cellres{2.2484}{0.04} & \cellres{2.0293}{0.04} & \cellres{0.4385}{0.02} & \cellres{0.4262}{0.02}
& \cellres{2.1227}{0.02} & \cellres{1.9084}{0.02} & \cellres{0.5126}{0.02} & \cellres{0.5028}{0.02} \\
\midrule
\multicolumn{10}{l}{\textbf{$\triangle$ Agent-designed methods}} \\
CellForge & arXiv'25
& \cellres{2.1038}{0.03} & \cellres{1.8937}{0.03} & \cellres{0.4978}{0.01} & \cellres{0.4896}{0.01}
& \cellres{1.9426}{0.02} & \cellres{1.7438}{0.02} & \cellres{0.5867}{0.03} & \cellres{0.5789}{0.03} \\
\textbf{CS-model} & Ours
& \cellbest{1.8127}{0.02} & \cellbest{1.6328}{0.01} & \cellbest{0.5924}{0.01} & \cellbest{0.5890}{0.01}
& \cellbest{1.7303}{0.02} & \cellbest{1.5517}{0.02} & \cellbest{0.6529}{0.03} & \cellbest{0.6518}{0.03} \\
\bottomrule
\end{tabular}
}
\end{table*}

\subsection{Single-cell Perturbation Prediction}
\label{app:res_trans}

Table~\ref{tab:singlecell_perturbation} reports task-specific workflow results on Norman and Papalexi, extending the evaluation beyond morphology and LINCS-style readouts. Norman evaluates transcriptional responses to gene activation; Papalexi evaluates RNA and protein responses as separate prediction tasks. Each comparison uses its dataset-specific protocol and reports PCC and $\mathrm{PCC}_{\mathrm{DE}}$.

\begin{table}[htpb]
\centering
\begin{minipage}[t]{0.85\textwidth}
\vspace{0pt}
\centering
\caption{
\textbf{Single-cell perturbation prediction.}
We compare task-specific CS-model predictors with CellForge on genetic and multimodal single-cell benchmarks using PCC and $\mathrm{PCC}_{\mathrm{DE}}$. Entries are point estimates under the dataset-specific protocols. Best values are \textbf{bolded}.
}
\label{tab:singlecell_perturbation}
\scriptsize
\renewcommand{\arraystretch}{0.94}
\setlength{\tabcolsep}{3pt}
\begin{tabular}{@{}llcc@{}}
\toprule
\textbf{Dataset} 
& \textbf{Method} 
& \textbf{PCC} $\uparrow$ 
& \textbf{$\mathrm{PCC}_{\mathrm{DE}}$} $\uparrow$ \\
\midrule

\multicolumn{4}{c}{\shortstack{\textit{CRISPRa Gene-Activation}\\\textit{scRNA-seq Dataset}}} \\
\midrule
\multirow{2}{*}{Norman}
& CellForge 
& 0.9846
& 0.8109 \\
& \textbf{CS-model}
& \textbf{0.9871}
& \textbf{0.9060} \\

\midrule
\multicolumn{4}{c}{\shortstack{\textit{CRISPR--Cas9 ECCITE-seq}\\\textit{RNA Dataset}}} \\
\midrule
\multirow{2}{*}{Papalexi RNA}
& CellForge 
& 0.6935
& 0.6406 \\
& \textbf{CS-model}
& \textbf{0.8193}
& \textbf{0.8311} \\

\midrule
\multicolumn{4}{c}{\shortstack{\textit{CRISPR--Cas9 ECCITE-seq}\\\textit{Protein Dataset}}} \\
\midrule
\multirow{2}{*}{Papalexi Protein}
& CellForge 
& 0.7495
& 0.7409 \\
& \textbf{CS-model}
& \textbf{0.7941}
& \textbf{0.8275} \\

\bottomrule
\end{tabular}
\end{minipage}
\end{table}

\end{document}

%% file: existing_run_statistics.tex
\subsection{Paired-run uncertainty}
\label{app:run_level_uncertainty}
For each fixed setting, $d_s=\mathrm{PCC}_{\mathrm{CS},s}-\mathrm{PCC}_{\mathrm{Random},s}$ pairs seeds $s\in\{11,22,33,44,55\}$. Table~\ref{tab:existing_controlled_paired_runs} reports $\bar d\pm s_d$ and the nominal interval $\bar d\pm t_{0.975,4}s_d/\sqrt{5}$, assuming independent approximately normal seed-level differences conditional on the fixed partition and candidate language. Table~\ref{tab:matched_pcc_effects} instead summarizes the four effects obtained after averaging seeds within each setting. The settings share two cohorts, budgets are repeated checkpoints of each run, and both backbone comparisons use the same Random-routing results. We therefore interpret the four-setting intervals descriptively and report seed-level uncertainty separately. Intervals are unadjusted across metrics and budgets. The accompanying CSVs retain all policy/checkpoint values and paired differences.
\begin{table}[!ht]
\centering
\footnotesize
\caption{\textbf{Paired-run CellScientist-minus-Random PCC differences.} Each cell gives mean$\pm$sample SD across five runs, with a nominal 95\% paired-$t$ interval below, conditional on the fixed task partition.}
\label{tab:existing_controlled_paired_runs}
\setlength{\tabcolsep}{3pt}
\renewcommand{\arraystretch}{1.35}
\begin{tabular}{@{}lllccc@{}}
\toprule
Backbone & Dataset & Split & $B=3$ & $B=5$ & $B=10$ \\
\midrule
Gemini 3 Pro & BBBC036 & Plate & \shortstack{$+0.0400\pm0.0491$\\{\scriptsize $[-0.0210,+0.1010]$}} & \shortstack{$+0.0397\pm0.0503$\\{\scriptsize $[-0.0228,+0.1022]$}} & \shortstack{$+0.0066\pm0.0067$\\{\scriptsize $[-0.0017,+0.0149]$}} \\
Gemini 3 Pro & BBBC036 & SMILES & \shortstack{$+0.0284\pm0.0406$\\{\scriptsize $[-0.0221,+0.0788]$}} & \shortstack{$+0.0297\pm0.0414$\\{\scriptsize $[-0.0217,+0.0811]$}} & \shortstack{$-0.0001\pm0.0019$\\{\scriptsize $[-0.0024,+0.0023]$}} \\
Gemini 3 Pro & BBBC047 & Plate & \shortstack{$+0.0364\pm0.0498$\\{\scriptsize $[-0.0254,+0.0983]$}} & \shortstack{$+0.0391\pm0.0514$\\{\scriptsize $[-0.0247,+0.1028]$}} & \shortstack{$+0.0023\pm0.0033$\\{\scriptsize $[-0.0017,+0.0064]$}} \\
Gemini 3 Pro & BBBC047 & SMILES & \shortstack{$+0.0398\pm0.0534$\\{\scriptsize $[-0.0265,+0.1061]$}} & \shortstack{$+0.0407\pm0.0583$\\{\scriptsize $[-0.0317,+0.1130]$}} & \shortstack{$+0.0083\pm0.0061$\\{\scriptsize $[+0.0007,+0.0159]$}} \\
\midrule
Qwen2.5-0.5B & BBBC036 & Plate & \shortstack{$+0.0316\pm0.0504$\\{\scriptsize $[-0.0309,+0.0942]$}} & \shortstack{$+0.0289\pm0.0510$\\{\scriptsize $[-0.0345,+0.0922]$}} & \shortstack{$+0.0031\pm0.0088$\\{\scriptsize $[-0.0078,+0.0140]$}} \\
Qwen2.5-0.5B & BBBC036 & SMILES & \shortstack{$+0.0304\pm0.0410$\\{\scriptsize $[-0.0205,+0.0813]$}} & \shortstack{$+0.0308\pm0.0415$\\{\scriptsize $[-0.0207,+0.0823]$}} & \shortstack{$-0.0001\pm0.0019$\\{\scriptsize $[-0.0024,+0.0023]$}} \\
Qwen2.5-0.5B & BBBC047 & Plate & \shortstack{$+0.0397\pm0.0501$\\{\scriptsize $[-0.0225,+0.1020]$}} & \shortstack{$+0.0359\pm0.0518$\\{\scriptsize $[-0.0285,+0.1002]$}} & \shortstack{$-0.0020\pm0.0039$\\{\scriptsize $[-0.0068,+0.0028]$}} \\
Qwen2.5-0.5B & BBBC047 & SMILES & \shortstack{$+0.0439\pm0.0534$\\{\scriptsize $[-0.0225,+0.1102]$}} & \shortstack{$+0.0407\pm0.0582$\\{\scriptsize $[-0.0315,+0.1130]$}} & \shortstack{$+0.0039\pm0.0075$\\{\scriptsize $[-0.0054,+0.0132]$}} \\
\bottomrule
\end{tabular}
\end{table}
For LKCP, seeds $1729$--$1733$ are matched by identifier on the same compound-disjoint test partition (16 compounds and 432 conditions). Table~\ref{tab:existing_lkcp_paired_runs} describes refit variation conditional on these compounds. Positive differences favor the CS-model for PCC and $R^2$; negative differences favor it for MSE. Seed matching preserves the run protocol, while architectures retain their own stochastic operations.
\begin{table}[!ht]
\centering
\footnotesize
\caption{\textbf{LKCP seed-matched CS-model-minus-RealMLP differences.} Mean$\pm$sample SD and nominal 95\% paired-$t$ intervals use five refits on the fixed test partition.}
\label{tab:existing_lkcp_paired_runs}
\setlength{\tabcolsep}{5pt}
\renewcommand{\arraystretch}{1.35}
\begin{tabular}{@{}lccc@{}}
\toprule
Frozen design & $\Delta$PCC & $\Delta$MSE & $\Delta R^2$ \\
\midrule
CS-model (BBBC036) & \shortstack{$-0.0046\pm0.0999$\\{\scriptsize $[-0.1286,+0.1194]$}} & \shortstack{$-0.0112\pm0.1604$\\{\scriptsize $[-0.2104,+0.1880]$}} & \shortstack{$+0.0006\pm0.0299$\\{\scriptsize $[-0.0365,+0.0377]$}} \\
CS-model (BBBC047) & \shortstack{$+0.0270\pm0.0672$\\{\scriptsize $[-0.0565,+0.1105]$}} & \shortstack{$+0.0583\pm0.1996$\\{\scriptsize $[-0.1896,+0.3061]$}} & \shortstack{$-0.0475\pm0.0564$\\{\scriptsize $[-0.1175,+0.0225]$}} \\
\bottomrule
\end{tabular}
\end{table}